\documentclass[final,5p,times,twocolumn,article]{elsarticle}

\usepackage{hyperref}       % hyperlinks
\hypersetup{
    colorlinks=true,
    linkcolor=blue,
    filecolor=magenta,      
    urlcolor=magenta,
    citecolor=blue,
}
\usepackage{float}    % allows for the [H] command
\usepackage{lipsum}
\usepackage{amssymb}
\usepackage{amsmath}
\usepackage{tabularx}
\usepackage{booktabs}
\usepackage{color, colortbl}
\usepackage{graphicx}
\usepackage{caption}
\usepackage{multirow}
\usepackage{subcaption}
\usepackage{xcolor}
\usepackage{paralist}
\usepackage{siunitx}
\definecolor{myblue}{rgb}{0.4, 0.6, 0.9}
\usepackage{longtable} % For tables that span multiple pages

\usepackage{lineno}

\journal{}

\begin{document}
% \setlength{\linenumbersep}{4pt}
% \linenumbers

\begin{frontmatter}



\title{Urban Safety Perception Assessments via Integrating Multimodal Large Language Models with Street View Images}

% \affiliation[label1]{organization={},
% \affiliation[label2]{organization={},

 \author[label1,label2]{Jiaxin Zhang}
 \ead{jiaxin.arch@ncu.edu.cn}
 \author[label1,label2]{Yunqin Li}
 \ead{Corresponding: liyunqin@ncu.edu.cn}
 \author[label2]{Tomohiro Fukuda}
 \ead{fukuda.tomohiro.see.eng@osaka-u.ac.jp}
 \author[label3]{Bowen Wang}
 \ead{wang@ids.osaka-u.ac.jp}

 \affiliation[label1]{organization={Architecture and Design College, Nanchang University},%Department and Organization
             addressline={No. 999, Xuefu Avenue, Honggutan New District}, 
             city={Nanchang},
             postcode={330031}, 
             country={China},
             }
 \affiliation[label2]{organization={Division of Sustainable Energy and Environmental Engineering, Osaka University},%Department and Organization
             addressline={2-1,Yamadaoka}, 
             city={Osaka},
             postcode={5650871}, 
             state={Suita},
             country={Japan},
             }
 \affiliation[label3]{organization={D3 Center, Osaka University},%Department and Organization
             addressline={2-1,Yamadaoka}, 
             city={Osaka},
             postcode={5650871}, 
             state={Suita},
             country={Japan},
             }

\begin{abstract}
Measuring urban safety perception is important and complex, traditionally relies heavily on resource-intensive field surveys, manual data collection, and subjective evaluations, which are costly and sometimes inconsistent. Street View Images (SVIs) combined with deep learning methods offer scalable urban safety detection but typically require extensive human annotation for model training, and differences in urban architecture hinder model transferability between cities. Therefore, a fully automated safety evaluation method is essential. Recent multimodal large language models (MLLMs), such as GPT, exhibit strong reasoning and analytical capabilities. We applied MLLMs to urban safety ranking using a human-annotated anchor dataset, validating their alignment with human perceptions. Additionally, we proposed a rapid city-wide safety assessment method leveraging pre-trained Contrastive Language-Image Pre-training (CLIP) features combined with K-Nearest Neighbors (K-NN) retrieval. In our experiments, GPT-4o achieves the closest alignment with human safety judgments ($R^2$ = 0.4031 in Chengdu and 0.4528 in Osaka). Our proposed CLIP+KNN also show superior prediction performance ($R^2$ = 0.4725 in Chengdu and 0.5112 in Osaka), surpassing existing state-of-the-art methods for large-scale urban safety evaluations. The proposed method provides an efficient, automated tool for large-scale urban safety evaluations, offering a reliable solution for researchers and practitioners aiming to improve urban environments.

\end{abstract}




\end{frontmatter}

%\tableofcontents


\section{Introduction}

Perceptions of safety in residential environments positively impact citizens' overall happiness \cite{pol2019safe,zhao2023}. Urban safety perception primarily emphasizes individuals' subjective feelings and psychological responses to the built environment in specific contexts \cite{qiu2023subjective}. Early researchers typically collected safety perception data through field surveys, on-site interviews, and questionnaires to assess the safety level of a given area \cite{wang2022measuring}. While these methods provide in-depth insights, they are often limited in scope, time-consuming, and costly. In contrast, new computer vision technology enables rapid, large-scale data acquisition and processing through automated image analysis, significantly improving efficiency and reducing costs \cite{xue2020extracting}. These approaches also allow researchers to capture safety perception data across various time points and environmental conditions, making studies more timely and adaptable.

In recent years, studies utilizing computer vision technology to uncover hidden city profiles through Street View Images (SVIs) have emerged to assess city-scale safety perceptions \cite{fan2023urban}. A primary challenge is the efficient acquisition of quantifiable data for training deep learning models to achieve robust model performance, given the subjective nature of safety perception \cite{yao2019human}. The development of MIT's Place Pulse project marked a significant shift in urban perception data collection methods \cite{salesses2012place}. By employing an online crowdsourcing strategy, participants compared pairs of images across six perceptual dimensions, resulting in the creation of the large-scale urban perception dataset Place Pulse 2.0 \cite{salesses2013collaborative}. Despite the demonstrated efficiency and effectiveness of deep learning models trained on Place Pulse 2.0 for predicting urban-scale safety perception scores \cite{zhang2018measuring}, the compilation of this dataset demanded considerable human effort, resources, and financial investment. Therefore, improving the efficiency of annotating subjective perception datasets and the transferability of models without increasing costs remains a key area for future efforts.

\begin{figure*}[t]
\centering
\includegraphics[width=1\linewidth]{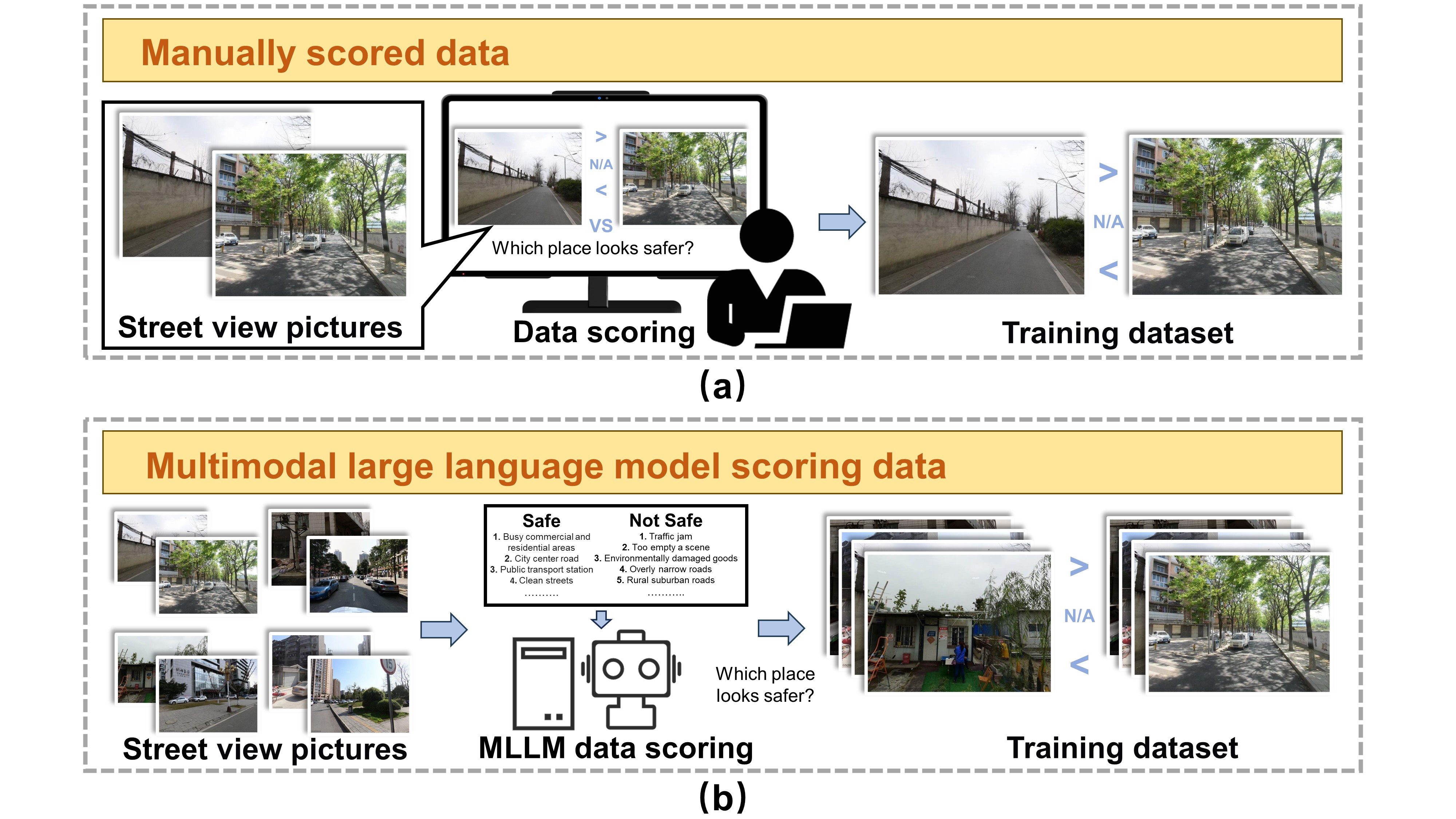}
\caption{Comparison of manually scored data and multimodal large language model scored data for urban safety perception. (a) Manually comparing pairs of SVIs for safety perception to create a training set, and (b) inputting two SVIs simultaneously into the MLLM to determine which image is perceived as safer.}
\label{fig: manual and machine}
\end{figure*}

Recently, large language models (LLMs) \cite{LLMSurvey} have demonstrated their proficiency in numerous sophisticated language understanding and generation tasks, significantly transforming the field of natural language processing. These models, which are trained on extensive text corpora, are adept at capturing the subtleties of human language, ranging from article writing and summarization to producing coherent and contextually appropriate text. This capability has been expanded to multimodal large language models (MLLMs) \cite{yin2023survey}, which have achieved remarkable successes. Models like GPT-4V \cite{gpt4} are now capable of not only understanding and generating text but also interpreting and analyzing visual information. This convergence of visual and linguistic processing allows them to respond to queries about image content, effectively narrowing the divide between visual perception and linguistic expression \cite{li2023blip,wang2023botcl}. By processing and comprehending complex visual scenes, these models can identify objects, detect patterns, and even deduce emotions or actions depicted in images \cite{liu2024visual,zhu2023minigpt, wang2023match}. It is interesting to explore the potential of MLLM as a substitute for human evaluation in scoring safety perception.

This research investigates the possibility of using MLLMs to automatically obtain safety scores from images. The study began by assembling two anchor sets of 1,000 SVIs from Chengdu, China and Osaka, Japan, respectively. Each image in the anchor set was annotated with a safety score derived from comprehensive human assessments. These scores serve as a benchmark for evaluating the efficacy of the proposed approach. These scores serve as a benchmark for evaluating the efficacy of our approach. MLLMs and predefined prompts are utilized to facilitate automatic scoring. Our goal is to use MLLMs to replace the safety comparison tasks traditionally performed by humans, mimicking the data-making process of the Place Pulse 2.0 dataset. Through this automated process, the model assigned a safety score to each SVI. A statistical analysis using the coefficient of determination ($R^2$) revealed a correlation of 0.3263 between the model-generated scores and human assessments. This finding suggests that MLLMs, particularly state-of-the-art models like GPT-4o \cite{gpt4}, can effectively undertake automatic scoring tasks.

Previous works on urban safety scoring rely heavily on large amounts of annotated data for training, which is both expensive and time-consuming to obtain, posing significant challenges for scalability and widespread implementation. We thus proposed a novel urban-wide safety scoring system that leverages a K-Nearest Neighbors (K-NN) retrieval method based on Contrastive Language-Image Pre-training (CLIP) \cite{radford2021learning} features. This innovative approach does not require a training phase, instead utilizing a weighted aggregation technique to enhance performance. Preliminary experimental results demonstrate that this method significantly outperforms traditional models that require extensive training, suggesting a promising direction for scalable, real-time urban safety assessments. Our findings not only underscore the capabilities of MLLMs in interpreting complex urban data but also pave the way for further advancements in automated SVIs-based safety analysis. 

Our contributions to this study can be distilled into three primary aspects: (1) A benchmark dataset is established for evaluating MLLMs in human urban perception, enabling standardized comparisons with human safety perception assessments across diverse city contexts. (2) We introduced MLLMs to analyze urban safety perception and compared the results with human perception outcomes, revealing no significant differences between the two. (3) An urban-wise scoring methodology utilizing CLIP features and K-NN retrieval is proposed, outperforming previous approaches that requires training.

\section{Related Works}
\subsection{Measuring Urban Safety Perception via SVIs}

Despite numerous efforts in urban safety research \cite{zhao2023, gehl2011life, yoshinobu1986aesthetic, jacobs1961death}, a clear and unified definition of perceived safety is lacking. Different studies define perceived safety in various ways \cite{Danish2024}. Mehta et al. define perceived safety as an individual's sense of security influenced by social and physical factors \cite{mehta2014evaluating}. Mouratidis et al. describe perceived safety as the level of comfort and risk perceived in the environment \cite{mouratidis2019impact}. Similarly, Qiu et al. emphasize that perceived safety is not an objective quantification but rather a subjective perception of the place \cite{qiu2023subjective}.

Some scholars equate perceived safety with fear of crime, but this is conceptually inaccurate \cite{zeng2022perceived}. In these studies, perceived safety is often used as an indicator to quantify urban crime rates \cite{makinde2020correlates}.  Zhang et al. show a mismatch between perceived safety and actual crime rates, and potential contradictions between urban spatial characteristics and ``perception bias" in crime \cite{zhang2021perception}. Kang et al. propose that the fear of crime is a trait reflecting individual differences in the experience of fear, whereas perceived safety is a situational and instantaneous state \cite{kang2023assessing}. In essence, perceived safety is a transient feeling influenced by the immediate perception of danger or threat, distinguishing it from the more enduring and pervasive nature of the fear of crime, which spans across various contexts and timeframes.

Field surveys, interviews, and questionnaires are the main methods for collecting perceived safety data \cite{wang2022measuring}. For instance, since 1973, the U.S. National Crime Survey (NCS) has collected data by asking: \textit{How safe do you feel, or would you feel safe walking alone in your neighborhood at night?} While these methods provide detailed insights into human perceptions and social background differences, they are limited by research scope, time-consuming data collection, and high costs \cite{zhang2018measuring}. When quantifying perceived safety, it is crucial to balance accuracy and scalability, ensuring subtle differences are captured while accommodating broader applications \cite{cui2023analysing}.

With the proliferation of SVIs and the widespread application of machine learning, crowdsourcing surveys using SVIs combine accuracy with scalability \cite{fan2023urban}. These images offer broad availability, ample sample sizes, and consistent spatial granularity, comprehensively representing the visual and morphological features of urban environments \cite{liu2023interpretable}. These methods not only provide the urban appearance in specific scenarios but also give volunteers instantaneous visual impressions, enabling the quantification of perception assessments \cite{cui2023analysing}. To better understand the role of perceived safety in urban environments, it is essential to consider quantification methods on a global scale. However, as noted by Kang et al. \cite{kang2023assessing}, differences between global models and local perceptions highlight the importance of contextualizing safety assessments to avoid embedding biases from prior studies. While manual statistical methods can capture individual and area-specific nuances, they are challenging to extend to systematic global studies. Therefore, combining prior knowledge of human safety perception with the capabilities of multimodal large language models may enable machine-supervised understanding, leading to advancements in broader applications and more precise quantitative analysis.

\subsection{Quantifying Urban Perception At a Large Scale}
Urban planners and sociologists have identified correlations between unsafe visual characteristics and issues such as crime and lower educational outcomes \cite{keizer2008spreading,kelling1982broken}. Nasar et al. \cite{nasar1990evaluative} posit that the visual and physical features of a city profoundly influence residents' cognition and emotional responses to urban spaces. However, quantitatively assessing perceptions of urban visual environments has been challenging due to limitations in data collection methods, such as insufficient sample sizes and reliance on interviews and surveys\cite{halpern2014mental,wang2024improving}. With advancements in image capture technology, SVIs have gradually become a significant medium for reflecting urban appearance.

Griew et al. developed the FASTVIEW audit tool to evaluate street environments using Google SVIs by assessing factors like pavement quality, lighting, and safety to determine their support for physical activity. The tool employs expert assessments or crowdsourced ratings, with participants scoring images from 1 to 10, where higher scores indicate environments more conducive to physical activity \cite{griew2013developing}. Salesses et al. created the Place Pulse 1.0 dataset by comparing pairs of SVIs to gather perceptions of urban environments. In this dataset, participants evaluated randomly selected image pairs, answering questions like \textit{Which place looks safer?} MIT's Place Pulse project marked a shift in urban perception data collection methods \cite{salesses2013collaborative}. Based on this dataset, Naik et al. \cite{naik2014streetscore} proposed the Streetscore algorithm, using support vector regression (SVR) and image features to predict perceived street safety. This algorithm analyzes features like texture, color, and shape to predict Streetscore, thereby automatically generating an urban appearance dataset for 21 U.S. cities \cite{naik2016cities}. However, because Streetscore was primarily trained on images from New York and Boston, its accuracy in measuring urban perception globally is limited.

Advances in deep learning for image recognition have driven the widespread application of convolutional neural network (CNN) models in urban visual quality perception studies based on the Place Pulse 1.0 dataset \cite{krizhevsky2012imagenet,porzi2015predicting,russakovsky2015imagenet}. Dubey et al. \cite{dubey2016deep} combined online crowdsourcing and deep learning techniques to create the Place Pulse 2.0 dataset, which includes 110,988 images and 1.17 million pairwise comparisons from 56 cities worldwide, rated by 81,630 volunteers. This dataset covers six perceptual dimensions (Safe, Lively, Beautiful, Wealthy, Depressing, and Boring). They also designed a Ranking Streetscore-CNN model to predict visual attributes of SVIs. These studies have not only improved data collection efficiency but also expanded the geographical coverage of samples, laying the foundation for global urban perception research.

Currently, the feasibility of machines replacing humans in rating SVIs is becoming a research focus. Naik et al. \cite{naik2017computer} used SVIs and the Place Pulse 2.0 dataset to study changes and drivers in visual quality perception of community appearances. However, the Place Pulse 2.0 dataset still has limitations in transferability, as deep learning models trained on this dataset perform poorly in predicting perceptions of SVIs from developing regions \cite{yao2019human}. Additionally, measuring models for street visual quality perception can be influenced by factors such as the age, gender, occupation, behavior, and the time of evaluation of the subjects \cite{cui2023analysing,fan2023urban,zhao2022,li2022measuring,liu2024day,lu2024using,wei2024measuring}.

\begin{figure*}[t]
\centering
\includegraphics[width=1\linewidth]{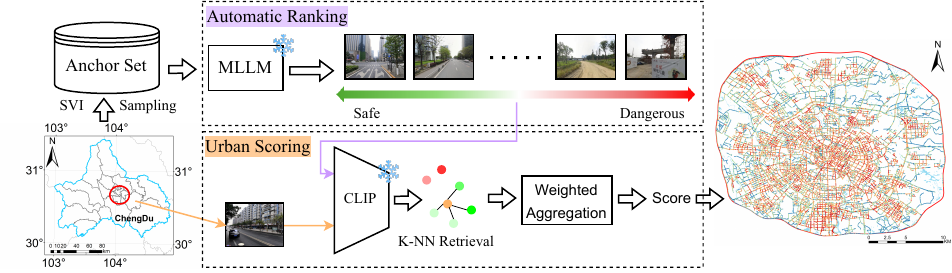}
\caption{The overall pipeline of our proposed method attaches an anchor set with safety perception scores via automatic ranking, followed by an urban scoring method to analyze the whole city.}
\label{fig:overview}
\end{figure*}

\subsection{Multimodal Large Language Models in Human Perception}
In recent years, the emergence of LLMs has brought about revolutionary changes in the field of Natural Language Processing (NLP), with models such as ChatGPT \cite{bahrini2023chatgpt}, GPT-4 \cite{achiam2023gpt}, PaLM \cite{chowdhery2023palm}, and LLaMa \cite{touvron2023llama} leading the charge. LLMs, owing to their vast corpora and intensive training computations, have demonstrated impressive capabilities in zero-shot and few-shot tasks, as well as in more complex tasks like mathematical problem-solving and common-sense reasoning. For instance, the advent of ChatGPT has highlighted the potential of LLMs in understanding human intent, reasoning, and following instructions to generate the required responses for specific tasks. Meanwhile, the introduction of GPT-4 \cite{gpt4} has unlocked tremendous potential for multimodal perception, which is crucial for real-world foundational capabilities.

Creating models for street view visual quality perception faces challenges of being time-consuming and potentially biased compared to human perception \cite{kang2023assessing}. The emergence of MLLMs offers new opportunities for developing universal models. Y. Zhang et al. \cite{zhang2023knowledge} proposed a purely visual approach that can generate textual descriptions of SVIs. Nevertheless, research on obtaining street visual quality perception evaluations through MLLMs via pairwise comparisons of SVIs remains absent. Rating methods may affect data consistency due to respondent habits and differences in understanding scales, while pairwise comparison methods are intuitive and reduce cognitive burden, suitable for decision-making scenarios with multiple options \cite{herbrich2006trueskill,louviere2015best,liu2023tcra}. MLLMs combine language understanding and image feature learning, offering the potential for comprehensive understanding and prediction of street view perceptions. This study aims to investigate whether MLLMs, pre-embedded with knowledge of safety perceptions in built environments, can emulate human evaluations of SVI safety. By doing so, it seeks to address the time-consuming nature of pairwise comparisons in image perception tasks and the transferability issues of pre-trained models.

\section{Methods}

\subsection{Overview of Our Method}
In this study, a novel approach was utilized to score the perceived visual safety of SVIs by leveraging a multimodal large language model (MLLM). Figure \ref{fig: manual and machine} shows the workflow of manually scored data and MLLM-scored data for urban safety perception. In previous methods, SVIs were manually scored by human annotators who compared pairs of images to determine which appeared safer. These manual scores formed the training dataset for the deep learning model. In our proposed method, the MLLM was employed to automatically score a new anchor set of SVIs. The MLLM, pre-trained on large and diverse datasets, inherently possesses general knowledge, including an understanding of common urban safety concepts. We did not fine-tune or train the MLLM using a custom dataset for this research. The automated scoring by the MLLM significantly expanded the scale of datasets that could be annotated, showcasing the model’s potential to replace humans for assessing perceptions of street safety efficiently. This approach not only ensures consistency but also minimizes the need for extensive manual labeling, making large-scale urban safety assessments more feasible.

The safety analysis process is illustrated in Figure \ref{fig:overview}, using Chengdu's central area as a case study. First, SVIs were collected from Baidu Maps, from which we randomly selected a representative anchor set. Human annotators then ranked this anchor set as the groundtruth to assess how closely the MLLM aligns with human judgment, referring to this set as the raw anchor set. Next, an MLLM assigned safety scores to these images in raw anchor set, producing a generated anchor set with updated scores for the same images. The next step, to evaluate the city's overall safety, we used a pre-trained CLIP model to extract features from both the generated anchor set images and all SVIs. A K-NN retrieval method, combined with a weighted aggregation technique, was used to compute the final scores. Finally, these scores were projected onto the city map to provide a comprehensive visual representation of urban safety.

\subsection{Dataset Making and Anchor Set} \label{data}
A total of 69,681 SVIs points from Chengdu were collected using the Baidu Maps API, and 45,378 SVIs points from Osaka were obtained using the Google Maps API for comparative experiments (each point has four directions 0°, 90°, 180°, and 270°), denoted as $\mathcal{X}=\{x_j^l\}$, where $j$ is the index of SVI and $l$ is the index of direction. From all the SVIs in Chengdu, we randomly sampled 1,000 representative images to form our anchor set $\mathcal{A}$. We then created a safety perception dataset of SVIs following the methodology used in the Place Pulse 2.0 dataset.

\begin{table}[t]
\caption{Basic safety criteria $C$ provided to participants for ranking tasks.}
\centering
\resizebox{1\columnwidth}{!}{
\begin{tabular}{l|l}
\toprule
Type & Descriptions \\
\midrule
Safe & Areas with high pedestrian activity, such as commercial buildings or residential zones \\
 & Public service facilities, including police stations and hospitals \\
 & Well-maintained and organized street trees \\
 & Sidewalks that are clean and in good repair \\
 & Active and clean downtown roads \\
 & Clearly marked and easily accessible public transport stops \\
 & Clear and visible road signs and directions \\
 & Well-maintained street decorations or public spaces \\
 & Presence of well-maintained greenery and parks\\
\midrule
Dangerous & Buildings that are damaged or abandoned\\
 & Walls that are blocked or in disrepair\\
 & Remote rural or suburban roads with little traffic\\
 & Areas where garbage is piled up or the environment is neglected\\
 & Active construction sites with insufficient safety measures\\
 & Areas lacking sufficient traffic lights\\
 & Complex and confusing traffic systems\\
 & High traffic areas with disorganized vehicle and pedestrian flow\\
 & Open land that is desolate and uninhabited\\
 & Narrow, enclosed spaces\\
 & Long and narrow roads or tunnels with poor visibility\\
\bottomrule
\end{tabular}}

\label{guide}
\end{table}

\textbf{Anchor Set}
We employed an online crowdsourcing method and invited 50 volunteers to randomly compare 1,000 representative SVIs in Chengdu to form our anchor set $\mathcal{A}$. The ages of the volunteers ranged from 18 to 65 years (mean age = 35), ensuring diversity in age, gender, and occupational backgrounds: 20\% were aged 18-25, 30\% aged 26-35, 25\% aged 36-45, 15\% aged 46-55, and 10\% aged 56-65. The gender distribution was balanced, with 52\% male and 48\% female participants. Occupational backgrounds included students (36\%), white-collar workers (10\%), freelancers (14\%), engineers (24\%), educators (10\%), and others (6\%). Each image is compared to 40 other images in average. We also implement the same annotation for Osaka.

Prior to the voting process, participants were provided with basic safety guidelines. As shown in Table \ref{guide}, these guidelines are derived from seminal works on street safety perception by Gehl \cite{gehl2011life}, Ashihara \cite{yoshinobu1986aesthetic}, and Jacobs \cite{jacobs1961death}, among others, as well as recent theoretical advancements in the field. A program was developed to sequentially sort images by assigning them numerical identifiers from 1 to 1000. The numbering reflects each image’s position within the Anchor Set but does not imply any subjective evaluation. For each image with identifier iii, it is compared against 40 other randomly selected images in pairwise evaluations. Participants are instructed to select the image they perceive as “safer” in response to a predetermined question \textit{Which place looks safer} or to indicate \textit{cannot compare} if they are unable to make a judgment, as illustrated in Figure 1(b). The mathematical expression for the safety perception score of each image is:

\begin{equation}
    S_i=\sum_{j=1}^{40}E(i,j)
\end{equation}

Where $S_i$ represents the final score for image $i$ , and  $E(i,j)$  denotes the evaluation outcome between image $i$ and the $j$-th randomly selected image. Specifically:

\begin{equation}
E(i, j) = 
\begin{cases} 
+1, & \text{if } i \text{ safer than} j \\ 
0, & \text{if can't compare } i \text{ and } j \\ 
-1, & \text{if } i \text{ dangerous than } j 
\end{cases}
\end{equation}

Each image i is compared against 40 randomly selected images, resulting in a cumulative score $S_i$ that serves as a quantitative measure of its safety perception. Utilizing this method, experiments were conducted with 50 participants, and the average score for each image was calculated based on these 50 annotations. Finally, the scores were normalized and multiplied by a factor of 10.

\subsection{Automatic Ranking}
As discussed earlier, the anchor set $\mathcal{A}$ functions as a prototype for scoring. The safety ranking within this set can represent the score distribution of SVIs across the entire city. The critical challenge here is determining how to assign a specific score to each image. One possible method, as outlined in Section \ref{data}, is through human voting. However, this approach is time-consuming, particularly when dealing with a large number of images. Additionally, human judgment is inherently subjective, necessitating a substantial number of participants to mitigate bias and achieve more objective results.

We thus proposed to use recent advanced MLLM to realize the scoring task automatically. Given two SVIs $x_1$ and $x_2$, the safety comparison is responded by:
\begin{equation}
    Response = MLLM(Q,x_1,x_2),
\end{equation}
where $Q$ represents the question formulated using a predefined prompt shown in Table \ref{table_sample_gpt4} (We also include criteria $C$ as a part of $Q$). In this prompt structure, we posit that the MLLM acts as an urban environment specialist, tasked with comparing the inputted SVIs and providing a brief rationale. Additionally, we supply the criteria $c$ employed in human scoring as a reference to guide the model's assessment. We show one sample of comparison using GPT-4V as MLLM in Table \ref{table_sample_gpt4}.

\begin{table}[t]
  \caption{A demonstration of comparing safety between two SVIs utilizing an MLLM, with GPT-4V as the current example. We begin by inputting two SVIs. Next, based on a predefined prompt, we engage the MLLM to assess and explain the relative safety of the two images, subsequently determining the safer option.}
  \centering
  \resizebox{1\columnwidth}{!}{
  \begin{tabular}{c m{8cm}}
    \hline
    \multicolumn{2}{c}{Visual Input Example for Safety Comparison} \\
    \hline
    \\
     & \multicolumn{1}{c}{\includegraphics[width=8cm,height=5.75cm,keepaspectratio]{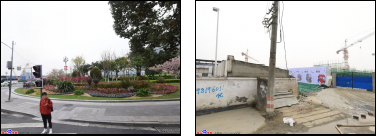}} \\
    \textbf{Question} & \multicolumn{1}{m{10cm}}{You are an urban environment expert. Here is the definition of safe and dangerous for city scenes: \{$C$\}. Now help me to compare the two input images and tell me which one is safer. Give me a choice from A: First Image or B: Second Image. C: Unable to compare. You also need to briefly explain your choice.} \\
    \hline
    \textbf{Response} & \multicolumn{1}{m{10cm}}{Choice: A: First Image. The first image depicts a well-maintained urban space with clear pedestrian pathways, neat rows of trees, and a vibrant garden area, which typically indicates a well-cared-for and safer environment. There are traffic signals in place and the presence of flowering plants suggests regular maintenance. Additionally, the area is open and visible, which can contribute to a sense of safety. In contrast, the second image shows a neglected area with debris, a lack of pedestrian infrastructure, and construction that appears unmanaged. The environment is less inviting, which may lead to a lower perceived level of safety. The presence of construction indicates potential hazards, and the overall disarray could be indicative of neglect. These elements typically correlate with a higher-risk environment and can contribute to a sense of unsafety.
} 
\\
    \hline
  \end{tabular}}
  \label{table_sample_gpt4}
\end{table}

Our automated ranking system is conceived to replicate the process of human voting. Each image $a_i \in \mathcal{A}$ is compared with $N=40$ images drawn from a randomly chosen subset $\mathcal{R}^i$ of $\mathcal{A}$ \footnote{Note that $a_i$ itself is not included in $\mathcal{R}^i$}. Furthermore, we maintain an index collection $\mathcal{S}$ for each $a_i$ to log the cumulative instances it is deemed safe, assigning 1 point per such designation. Hence, the ranking is denoted by the following equation:
\begin{equation}
    \mathcal{S} = \sum_{i=1}^I \sum_{n=1}^N \Theta(MLLM(Q, a_i, r^i_n)),
\end{equation}
where $\sum$ is the loop operation, $I$ symbolizes the total count of images in $\mathcal{A}$, $r^n_i$ is one image from $\mathcal{R}^i$ and $\Theta$ represents the tallying function that records an image as safe as determined by the MLLM. Note that if the response is ``Unable to compare" the result will not be counted.

After the ranking process, we obtain $\mathcal{S}$ that encapsulates the safety score $s_i$ for each $a_i$. To facilitate comparison and interpretation, it is imperative to normalize these scores to a uniform range. We accomplish this by mapping the scores to a scale from 0 to 10, which enables a standardized evaluation of safety across different images. The normalization formula is given by:
\begin{equation} \label{qwwww}
    \mathcal{S}^* = 10\frac{\mathcal{S} - min(\mathcal{S})}{max(\mathcal{S}) - min(\mathcal{S})}, 
\end{equation}
where $max$ and $min$ denote the operations to extract the highest and lowest scores within the set $\mathcal{S}$, respectively. This standardization process not only renders the scores more intuitive by placing them on a decile scale but also mitigates the effects of outliers, thereby providing a more accurate and robust comparison of safety levels.

\subsection{Urban Scoring Via CLIP and K-NN Retrieval}
Given the extensive number of SVIs contained within $\mathcal{X}$, directly appraising the safety of an entire city via the ranking method delineated in the previous section is an infeasible task. A common approach to overcome this challenge would be to develop a dedicated scoring deep learning model \cite{salesses2013collaborative} that processes individual SVIs and outputs corresponding safety scores based on the anchor set. Nonetheless, the development of such a model is fraught with difficulties, compounded by the potential for inaccuracies, particularly when the available data is sparse.

To address these problems, we propose an innovative solution that leverages a pre-trained CLIP model in conjunction with a K-NN retrieval system. CLIP is a model developed by OpenAI that learns multimodal representations through contrastive learning from a large number of image-text pairs (400 million). Specifically, it embeds images and texts into a shared vector space and trains by maximizing the similarity between matching pairs. This allows CLIP to serve as a powerful backbone for image feature extraction, converting images into numerical vector representations for downstream tasks, such as classification or retrieval. Our approach harnesses the robust feature extraction capabilities of CLIP to analyze the visual content of SVIs and understand their contextual nuances in relation to safety. By integrating these features with a K-NN retrieval mechanism, we can effectively match each city's SVI with its closest counterparts in the anchor set. This process not only bypasses the intensive data requirements of training a bespoke model from scratch but also enhances the accuracy of the safety scores through the use of a sophisticated, pre-established knowledge base. Thus, we can extrapolate a comprehensive safety score for the city, offering a nuanced and data-rich safety landscape.

Here, we represent the pre-trained CLIP model as $M$. The feature extraction process for all SVIs within the set $\mathcal{A}$ is mathematically expressed as follows:
\begin{equation}
\mathcal{F} = \{M(a_i) \mid a_i \in \mathcal{A}\},
\end{equation}
where $\mathcal{F}$ is the collection of feature vectors, and each $f_i$ is a vector extracted from the last layer of the CLIP image encoder.

For any given SVI $x^l_j$ from $\mathcal{X}$, we denote its extracted feature vector by $q^l_j$. The similarity measurement between $q^l_j$ and each anchor feature $f_i$ is computed using the cosine similarity metric $sim$, resulting in the feature distance:
\begin{equation}
\mathcal{D}^l_j = \{sim(q^l_j, f_i) \mid f_i \in \mathcal{F}\},
\end{equation}
where $\mathcal{D}^l_j$ encapsulates the similarity between $x^l_j$ and each SVI in the anchor set.

\begin{table*}[t]
	\caption{The score difference between human vote and automatic ranking via MLLM. Mean Absolute Error (MAE) is computed for the score differences relative to each SVI in the anchor set, with ± Std. indicating the standard deviation. Max and min are the maximum and minimum difference values, respectively. R$^2$ is the coefficient of determination.}
	\label{table_ranking_eval}
	\centering
	\resizebox{0.8\textwidth}{!}{%
    	\begin{tabular}{lcccccccc}
    		\toprule
    	    \multirow{2}{*}{Methods} & \multicolumn{4}{c}{Chengdu} & \multicolumn{4}{c}{Osaka}\\
            \cmidrule(lr){2-5} \cmidrule(lr){6-9}
    	    &MAE $\downarrow$ &  Max $\downarrow$ & Min $\downarrow$ & R$^2$ $\uparrow$ &MAE $\downarrow$  &  Max $\downarrow$ & Min $\downarrow$ & R$^2$ $\uparrow$\\
    	    \midrule
            \rowcolor{blue!20} 
            GPT-4o & 0.5872 ± 0.4735 & 2.8117 & 0.0014 & 0.4031 & 0.5038 ± 0.4235 & 2.3149 & 0.0027 & 0.4528 \\
            GPT-4V & 0.9157 ± 0.6916 & 3.7605 & 0.0027 & 0.3263 & 0.8214 ± 0.6139 & 3.3293 & 0.0025 & 0.3575 \\
            LLaVA-NeXT & 1.0514 ± 0.7332 & 3.9662 & 0.0035 & 0.2637 & 1.0332 ± 0.7187 & 3.4517 & 0.0041 & 0.2589 \\
            Phi3-V & 1.1902 ± 0.8099 & 4.1321 & 0.0032 & 0.2003 & 1.0298 ± 0.7165 & 3.7092 & 0.0030 & 0.2485 \\
            LLaMA-Adapter & 1.2377 ± 0.8386 & 4.6744 & 0.0031 & 0.1817  & 1.0951 ± 0.7614 & 3.8743 & 0.0034 & 0.2216 \\
            InstructBLIP & 1.3328 ± 0.9031 & 4.8947 & 0.0039 & 0.1452  & 1.3035 ± 0.8697 & 4.6204 & 0.0044 & 0.1557 \\
            MiniGPT-4 & 1.3501 ± 0.9178 & 5.0100 & 0.0046 & 0.1319  & 1.4167 ± 0.9324 & 5.2048 & 0.0050 & 0.1202 \\
    		\bottomrule
    	\end{tabular}
	}
\end{table*}

To derive a weighted relevance score for $x^l_j$, we aggregate the scores of its top K-nearest neighbors. This aggregation is performed by weighting the score of each neighbor by its relative distance, which emphasizes closer neighbors and potentially increases the reliability of the relevance score:
\begin{equation}
o^l_j = \sum_{k=1}^K s_k \left(\frac{d_k}{\sum_{k=1}^K d_k}\right),
\end{equation}
where $s_k$ is the recorded score of the k-th neighbor in the ranking list $\mathcal{S}^*$ of anchor set (defined in Equation \ref{qwwww}), and $d_k$ represents the distance between $x^l_j$ and its k-th nearest neighbor. The normalization by the sum of distances ensures that the weighted scores sum to one, maintaining a probabilistic interpretation of the relevance scores. This methodological framework allows for nuanced insights into the relationships between images based on deep feature similarities, providing a robust result for image-based retrieval scoring.

The final safety score for data point $j$ is calculated by averaging the scores from four SVIs associated with it, as follows:
\begin{equation}
    o_j = \frac{1}{4}\sum_l^4o^l_j
    .
\end{equation}
After calculating the safety scores for all data points, a non-linear analysis of SVI visual indicators is conducted, followed by hierarchical clustering of the local effects of these variables. This process results in a comprehensive safety map that reveals the complex relationships between visual elements and perceived safety scores, identifying critical thresholds where these elements transition from enhancing to diminishing safety perception. The integration of non-linear analysis and mapping provides a systematic understanding of the relationship between CLIP-derived features and safety perception, effectively validating the model’s capability to capture human perceptions of urban safety.

\section{Results}
\subsection{Experimental Settings}
In selecting the MLLMs, we have incorporated five recent state-of-the-art (SOTA) open-source methods: LLaVA-NeXT \cite{liu2024llavanext}, Phi3-V \cite{abdin2024phi}, LlaMA-Adapter \cite{zhang2023llama}, InstructBLIP \cite{dai2024instructblip}, and MiniGPT-4 \cite{zhu2023minigpt}. We also include advanced close-source models GPT-4V and GPT-4o in our experiments. The temperature value is set as 0.05 for all the MLLMs. Continuously, we evaluate our K-NN retrieval approach using the regenerated anchor set by GPT-4o. For the pre-trained CLIP model, we employ a version equipped with a ViT-B/16 backbone, which outputs a 512-dimensional feature vector. The official weight for OpenAI is used and no finetuning is implemented. 80\% of the data in the generated anchor set are used for training and the rest for evaluation. The K-NN retrieval is set with a default maximum $K$ value of 10. All experiments are conducted on a GPU server equipped with four Nvidia A40 GPUs.

Following \cite{salesses2013collaborative}, we adopted the coefficient of determination $R^2$ to evaluate the difference between the two results, which is formulated as follows:
\begin{align}
    SST = \sum^T_{t=0}(y_t-\bar{y})^2, \\
    SSE = \sum^T_{t=0}(y_t-\hat{y}_t)^2, \\
    R^2 = 1 - \frac{SSE}{SST},
\end{align}
where $y_t$ are the observed values (from the human vote), $\bar{y}$ is the mean of the dependent variable, and $\hat{y}_t$ are the values predicted by the model (MLLMs or CLIP based retrieval). We also adopt Mean Absolute Error (MAE) for further evaluation.
\begin{figure}[t]
\centering
\includegraphics[width=1\linewidth]{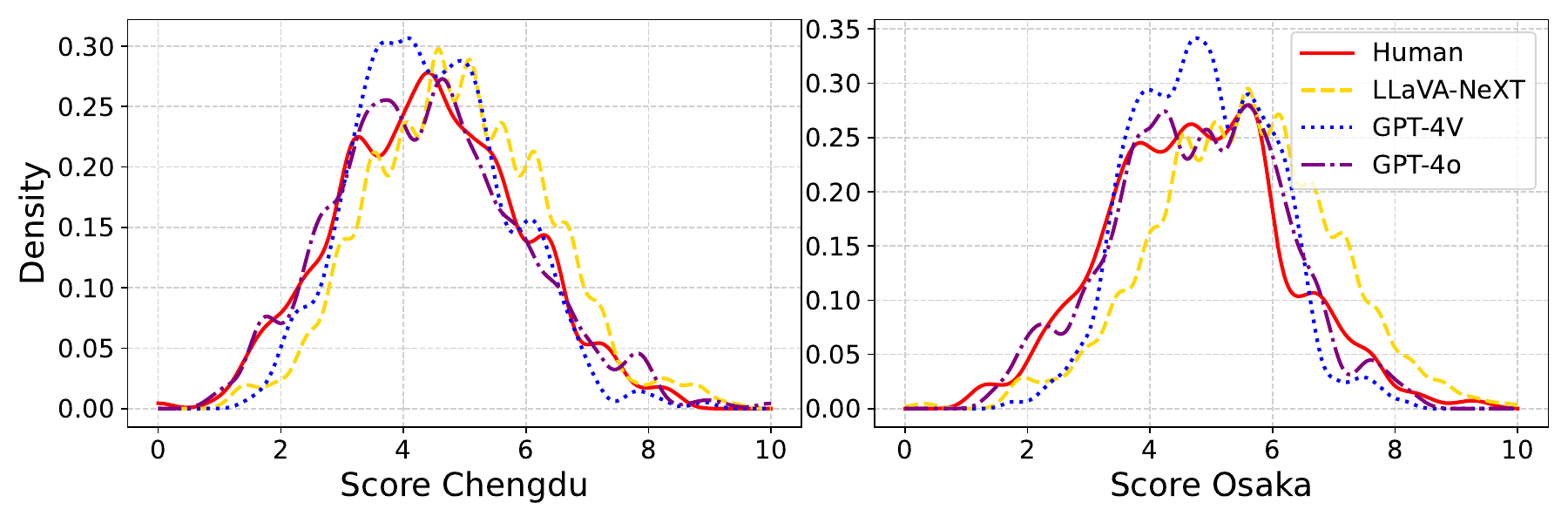}
\caption{Probability density of scores for the anchor set calculated from human votes and MLLM automatic rankings (LLaVA-NeXT, GPT-4V and GPT-4o), with the x-axis representing scores from 0 to 10 and the y-axis showing the distribution density.}
\label{fig:ranking}
\end{figure}

\begin{figure*}[t]
\centering
\includegraphics[width=1\linewidth]{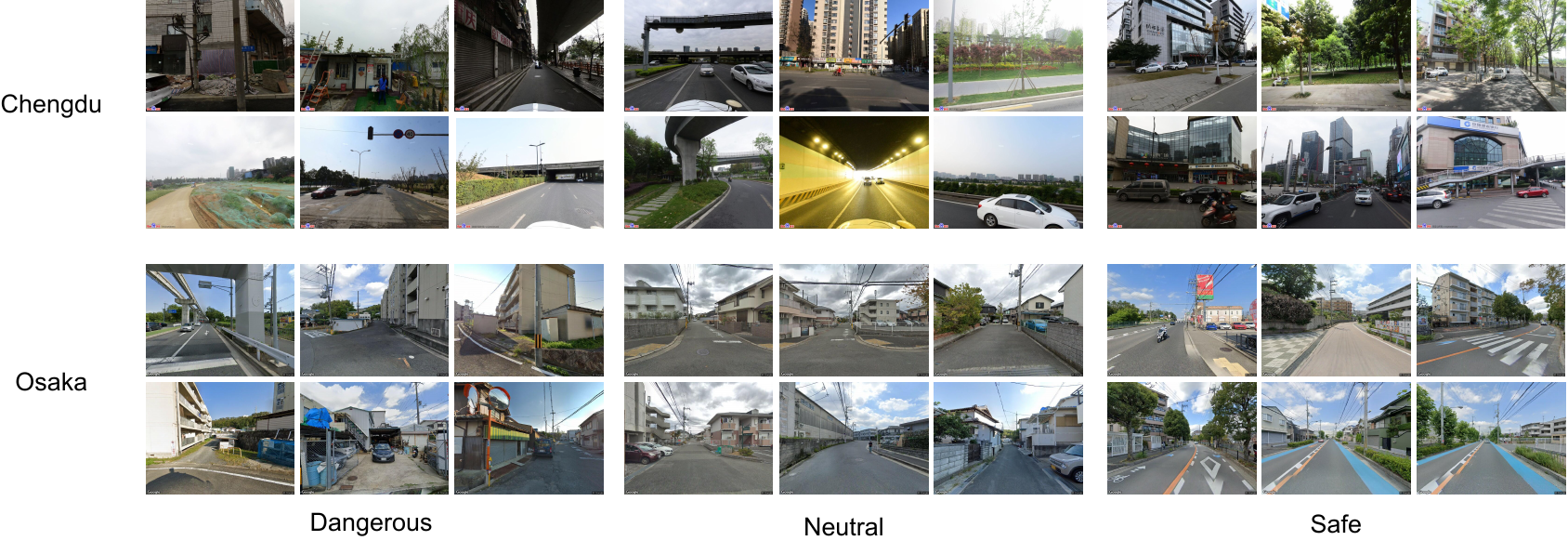}
\caption{Depicted here are examples of various types of SVIs from Chengdu city and Osaka city, categorized based on their assigned risk scores. From left to right, the classifications are as follows: \texttt{Dangerous}, indicated by a score less than 3; \texttt{Neutral}, characterized by scores ranging from 3 to 6; and \texttt{Safe}, denoted by scores exceeding 6. Each category reflects the relative safety level, with \texttt{Dangerous} representing high-risk conditions, \texttt{Neutral} indicating moderate risk, and \texttt{Safe} signifying low or negligible risk. }
\label{fig:samples}
\end{figure*}

\begin{figure}[t]
\centering
\includegraphics[width=0.8\linewidth]{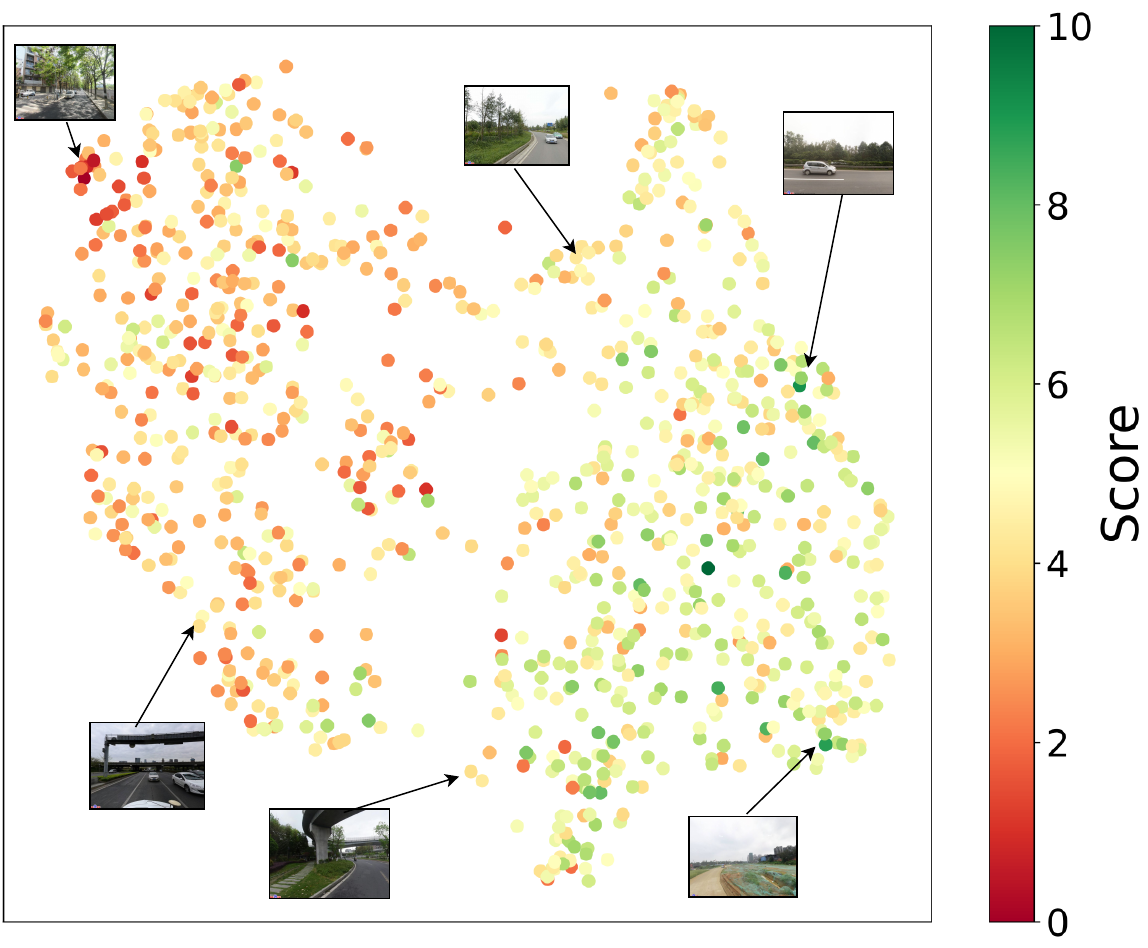}
\caption{UMAP visualization of the anchor set's image features extracted by CLIP, with safety scores represented by a color gradient from red (dangerous) to green (safe) and accompanied by sample images for reference.}
\label{fig:umap}
\end{figure}

\subsection{Evaluation of Automatic Ranking} \label{auto}
The primary objective of this research is to investigate the capability of MLLMs to align with human judgments in the task of urban safety ranking. This alignment is crucial for developing reliable AI tools that can assist in urban planning and public safety assessments. 

As shown in Table \ref{table_ranking_eval}, we present a comparative analysis of the performance of several leading MLLMs. Notably, GPT-4o excels across multiple key metrics, surpassing other models. For Chengdu, GPT-4o achieves the lowest Mean Absolute Error (MAE) (Mean ± Std) at 0.5872 ± 0.4735, reflecting the smallest discrepancy between its scores and human ratings, as well as greater stability with minimal impact from extreme values. Its maximum error (Max) is 2.8117, the lowest among all models, indicating a narrower error range across instances. Additionally, GPT-4o's R$^2$ is 0.4031, significantly outperforming other models in its ability to explain data variability and provide a better fit\footnote{Considering the inherent subjectivity of human evaluations, we consider this R² score sufficient to demonstrate GPT-4o’s alignment with human judgment in urban safety ranking tasks.}. In Osaka, GPT-4o similarly demonstrates superior alignment with human scoring, achieving an R$^2$ of 0.4528. Moreover, GPT-4o shows a notable advantage over GPT-4V, with improvements of 0.0768 and 0.0943 in R$^2$ for Chengdu and Osaka, respectively. This suggests that employing advanced models can reduce the gap between MLLMs and human assessments. We believe these results further highlight the potential of MLLMs in real-world applications, providing a stronger foundation for future research.

% LLava-NeXT has a MAE of 1.0514 ± 0.7332, which, while relatively good among the models, has a higher maximum error of 4.0662, a minimum error of 0.0038, and an R$^2$ of 0.2637, indicating larger errors in some instances and limited ability to explain data variability. LLaMA-Adapter's MAE is 1.2377 ± 0.8386, with a maximum error of 4.6744, a minimum error of 0.0031, and an R$^2$ of 0.1817, showing poorer prediction accuracy, larger errors, and inadequate fitting. InstructBLIP's MAE is 1.3328 ± 0.9031, one of the highest among the models, with a maximum error of 4.8947, a minimum error of 0.0039, and an R$^2$ of 0.1452, indicating a wide error range and the poorest ability to explain data variability. MiniGPT-4 has the highest MAE at 1.3501 ± 0.9178, with a maximum error of 5.0100, a minimum error of 0.0046, and an R$^2$ of 0.1319, the lowest among the models, indicating the poorest prediction accuracy, the highest errors, and the worst fit. 

Figure \ref{fig:ranking} illustrates the probability density of scores for the generated anchor set, with results for Chengdu on the left and Osaka on the right. These scores are derived from both human votes and automatic rankings by MLLMs, including LLava-NeXT, GPT-4V, and GPT-4o. The x-axis represents the score range from 0 to 10, while the y-axis indicates the distribution density. In Chengdu's case, the probability density curve for human scores, depicted by the solid red line, is relatively smooth, with a notable density peak around 4. This suggests that human scores are predominantly concentrated within this range. The blue dotted line, representing GPT-4o, aligns closely with human scores, also peaking around a score of 4. In contrast, GPT-4V displays some density differences, suggesting a more neutral distribution that does not closely align with human evaluations. The yellow dashed line for LLava-NeXT is skewed to the right, indicating that its scores tend to be higher and diverge from human scores within this range. The Osaka results reveal a similar pattern, with human scores displaying a smooth distribution across the 4 to 6 range, suggesting a slightly broader concentration compared to Chengdu.

\subsection{Evaluation for Scoring}
Our analysis reveals that CLIP serves as a highly effective backbone for extracting image features. Rather than training a separate classifier, we leverage these features directly to calculate the similarity between SVIs. This approach not only simplifies the process but also enhances the efficiency and accuracy of scoring measurements. 

As shown in Figure \ref{fig:umap}, as proof, we showcase the results of uniform manifold approximation and projection (UMAP) \cite{mcinnes2018umap} visualization of image features extracted using CLIP. Each point represents an image, with its safety score indicated by a color gradient ranging from red to green, where red denotes danger and green indicates safety. The distinct clusters formed by the points of different colors demonstrate that CLIP effectively captures image features, grouping similar image features together. The clear distribution of safety scores, as shown by the color gradient, indicates that the features extracted by CLIP can accurately reflect differences in image safety. UMAP brings together image points with similar features, suggesting that the features extracted by CLIP have a good distinguishing capability and can effectively reflect the similarity between images. This supports the choice of directly using CLIP features for similarity calculation, rather than training a classifier, making the method both efficient and accurate for scenarios requiring quick assessment of image similarity. Some samples are shown in Figure \ref{fig:samples}.

Additionally, in Table \ref{table_dual}, we present the evaluation results for various methods applied to safety score prediction. Our proposed method is highlighted with an asterisk (*). The baseline models, including ResNet-50, ViT-B/16, CLIP (RN50), and CLIP (ViT-B), were fine-tuned with a regression head for score prediction. We also incorporate four recent regression methods in our comparisons: Streetscore \cite{naik2014streetscore}, a classification approach leveraging a combination of different features; Adaptive Contrast \cite{dai2021adaptive}, a contrastive learning method designed specifically for medical image regression tasks; UCVME \cite{dai2023semi}, a semi-supervised approach for regression; and RESupCon \cite{RESupCon}, a supervised contrastive learning technique. The performance of each method is assessed using the R$^2$ values and Mean Error metrics. For Chengdu, ResNet-50 and ViT-B/16 (both pre-trained on ImageNet) show R$^2$ values of 0.1288 and 0.1455, respectively, indicating limited prediction capability for safety scores. CLIP models (RN50 and ViT-B) achieve R$^2$ values of 0.3805 and 0.4364, markedly outperforming ImageNet-pre-trained models, thus demonstrating better capacity for feature extraction and safety score prediction. Among the recent regression methods, only RESupCon slightly surpasses CLIP (ViT-B) in R$^2$ performance. Our proposed CLIP+KNN approach achieves an R$^2$ value of 0.4725, the highest among all methods. This approach leverages CLIP-extracted features directly, combining them with KNN for score prediction without requiring an additional regression head. These results demonstrate the effectiveness and advantage of our approach. We observe a similar performance pattern in our experiments for Osaka, where our method consistently achieves the best results.

\begin{table}[t]
	\caption{Evaluation on different methods for safety score prediction. Our method is attached with *.}
	\label{table_dual}
	\centering
	\resizebox{1\columnwidth}{!}{%
    	\begin{tabular}{lcccc}
    		\toprule
    	      \multirow{2}{*}{Methods} & \multicolumn{2}{c}{Chengdu} & \multicolumn{2}{c}{Osaka}\\
            \cmidrule(lr){2-3} \cmidrule(lr){4-5}
    	    &\textcolor{blue}{MAE} $\downarrow$  & R$^2$ $\uparrow$ &\textcolor{blue}{MAE} $\downarrow$  & R$^2$ $\uparrow$\\
    	    \midrule
            ResNet-50 & 1.3754 ± 0.9025  & 0.1288 & 1.2903 ± 0.8540 & 0.1502 \\
            ViT-B/16 & 1.2458 ± 0.8742 & 0.1455 & 1.1701 ± 0.8105 & 0.1780 \\
            CLIP (RN50) & 0.7850 ± 0.6200 & 0.3805 & 0.7503 ± 0.6012 & 0.3958 \\
            CLIP (ViT-B) & 0.5703 ± 0.4631 & 0.4364 & 0.5002 ± 0.4124 & 0.4950 \\
            \midrule
            Streetscore \cite{naik2014streetscore} & 0.9603 ± 0.7134 & 0.3170 & 1.0352 ± 0.7389 & 0.2905 \\
            Adaptive Contrast \cite{dai2021adaptive} & 1.0535 ± 0.7818 & 0.2847 & 0.9981 ± 0.7402 & 0.3103 \\
            UCVME \cite{dai2023semi} & 0.8027 ± 0.6615 & 0.3605 & 0.7884 ± 0.6508 & 0.3720 \\
            RESupCon \cite{RESupCon} & 0.5510 ± 0.4473 & 0.4428 & 0.4989 ± 0.4083 & 0.4975 \\
            \midrule
            \rowcolor{blue!20} 
            CLIP+KNN* & 0.5290 ± 0.4135 & 0.4725 & 0.4813 ± 0.4023 & 0.5112  \\
    		\bottomrule
    	\end{tabular}
	}
\end{table}

\begin{figure}[t]
\centering
\includegraphics[width=0.95\columnwidth]{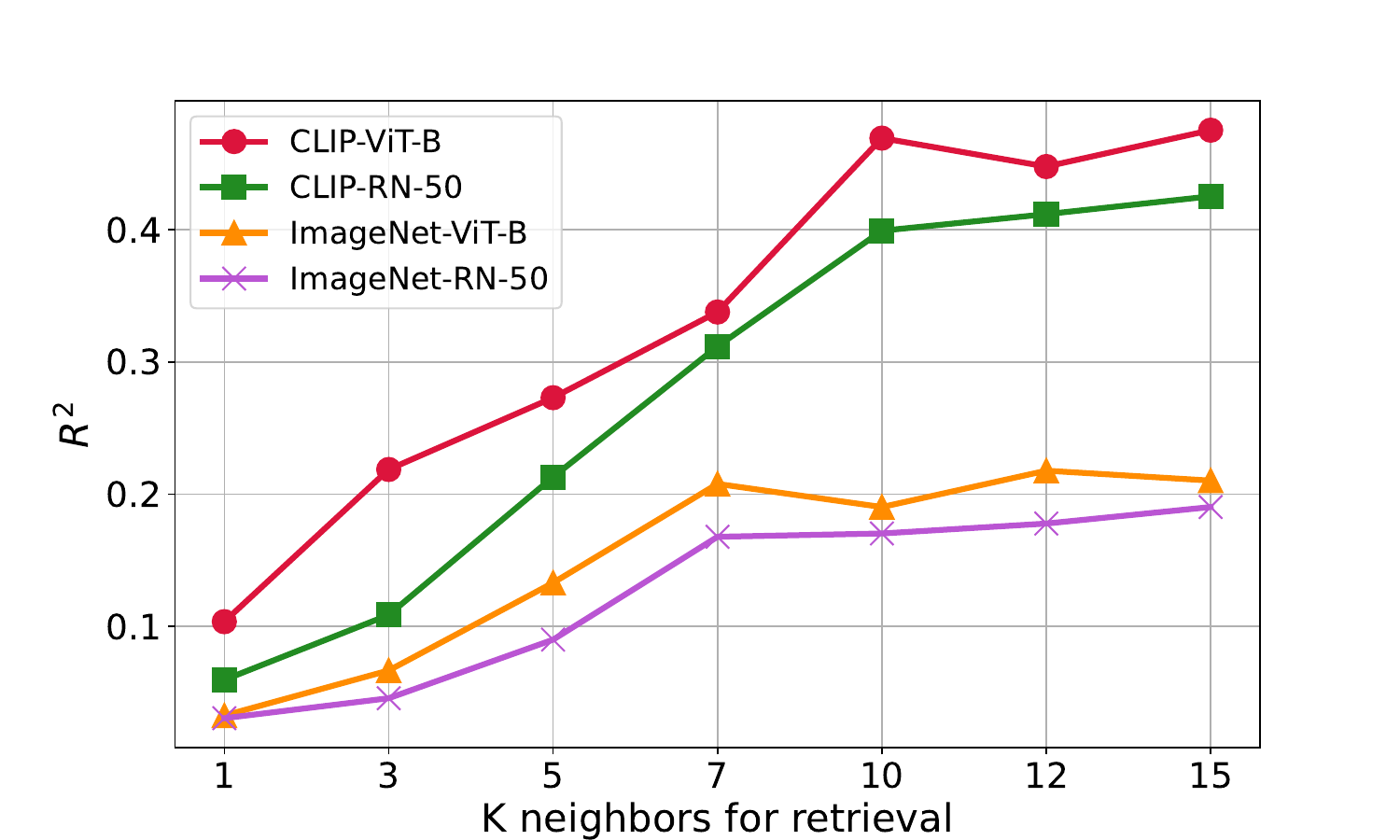}
\caption{Analysis of using top K neighbors for our K-NN based retrieval, comparing various backbone architectures with different pre-training patterns, including CLIP and ImageNet. Experiments implemented on generated Chengdu anchor set using GPT-4o.}
\label{fig:retrieval}
\end{figure}

\begin{figure}[t]
\centering
\includegraphics[width=1\columnwidth]{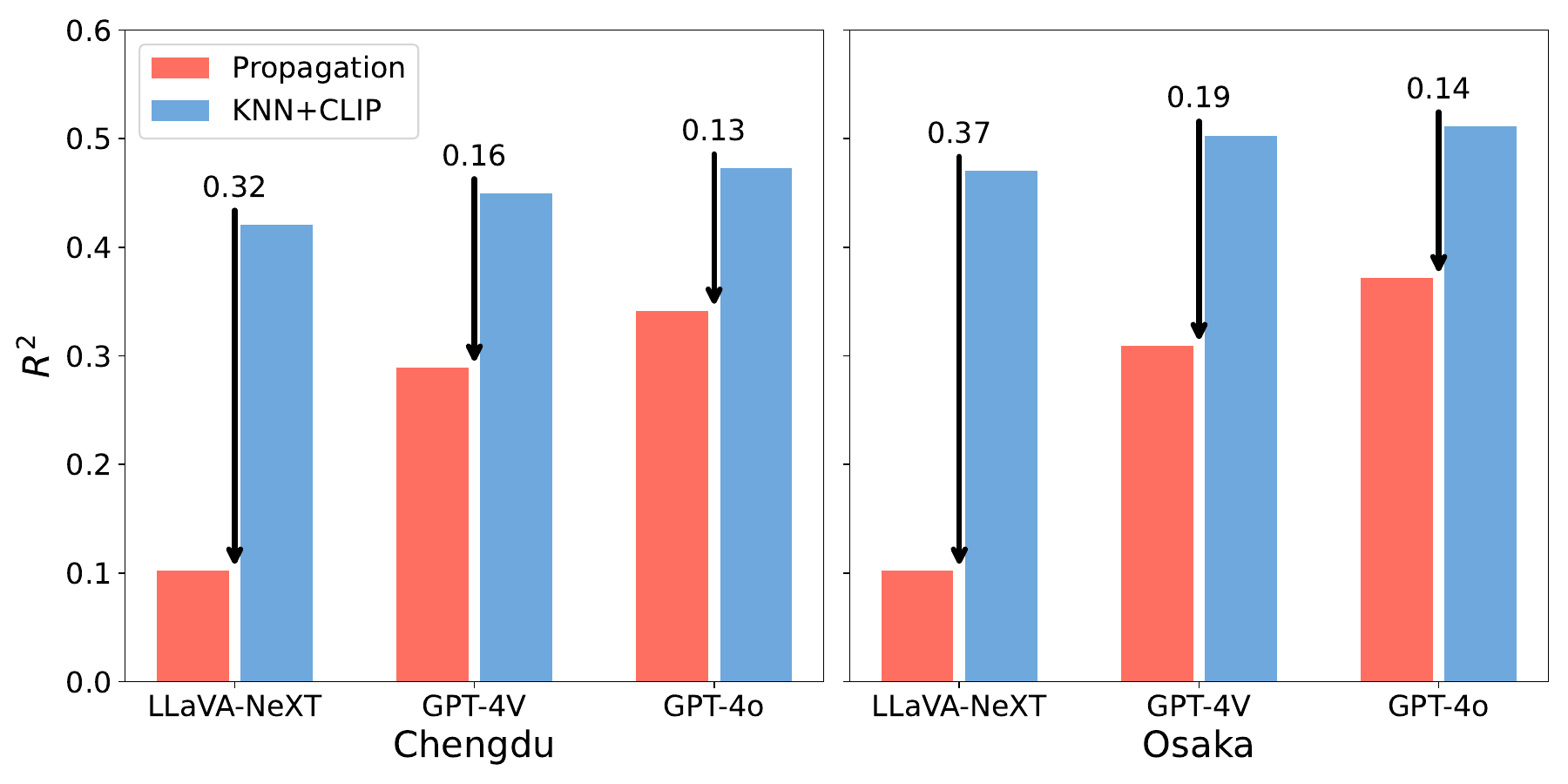}
\caption{Error propagation experiments. We use LLaVA-NeXT, GPT-4V, and GPT-4o to generate the anchor set for continues KNN+CLIP, respectively.}
\label{fig:propagation}
\end{figure}

We also provided an ablation study for the choice of $K$. Figure \ref{fig:retrieval} illustrates the R² performance of various backbone architectures using different numbers of neighbors ($K$ values) for K-NN based retrieval. It compares the performance of architectures pre-trained with CLIP (CLIP-ViT-B and CLIP-RN-50) and ImageNet (ImageNet-ViT-B and ImageNet-RN-50). Among them, CLIP-ViT-B consistently performs the best across all $K$ values, with its R$^2$ value steadily increasing as $K$ increases from 1 to 10, peaking at 0.4725, and then slightly decreasing but remaining high. CLIP-RN-50 also shows improvement with increasing $K$ values. In contrast, both ImageNet-ViT-B and ImageNet-RN-50 perform significantly worse across all $K$ values, with lower R$^2$ values indicating weaker feature extraction and similarity calculation capabilities. Overall, the result demonstrates that the general trend of increasing R$^2$ values with larger $K$ values suggests that using more neighbors improves prediction accuracy, although the benefits level off or slightly decreases beyond a certain point.

\begin{figure*}[t]
\centering
\includegraphics[width=1\textwidth]{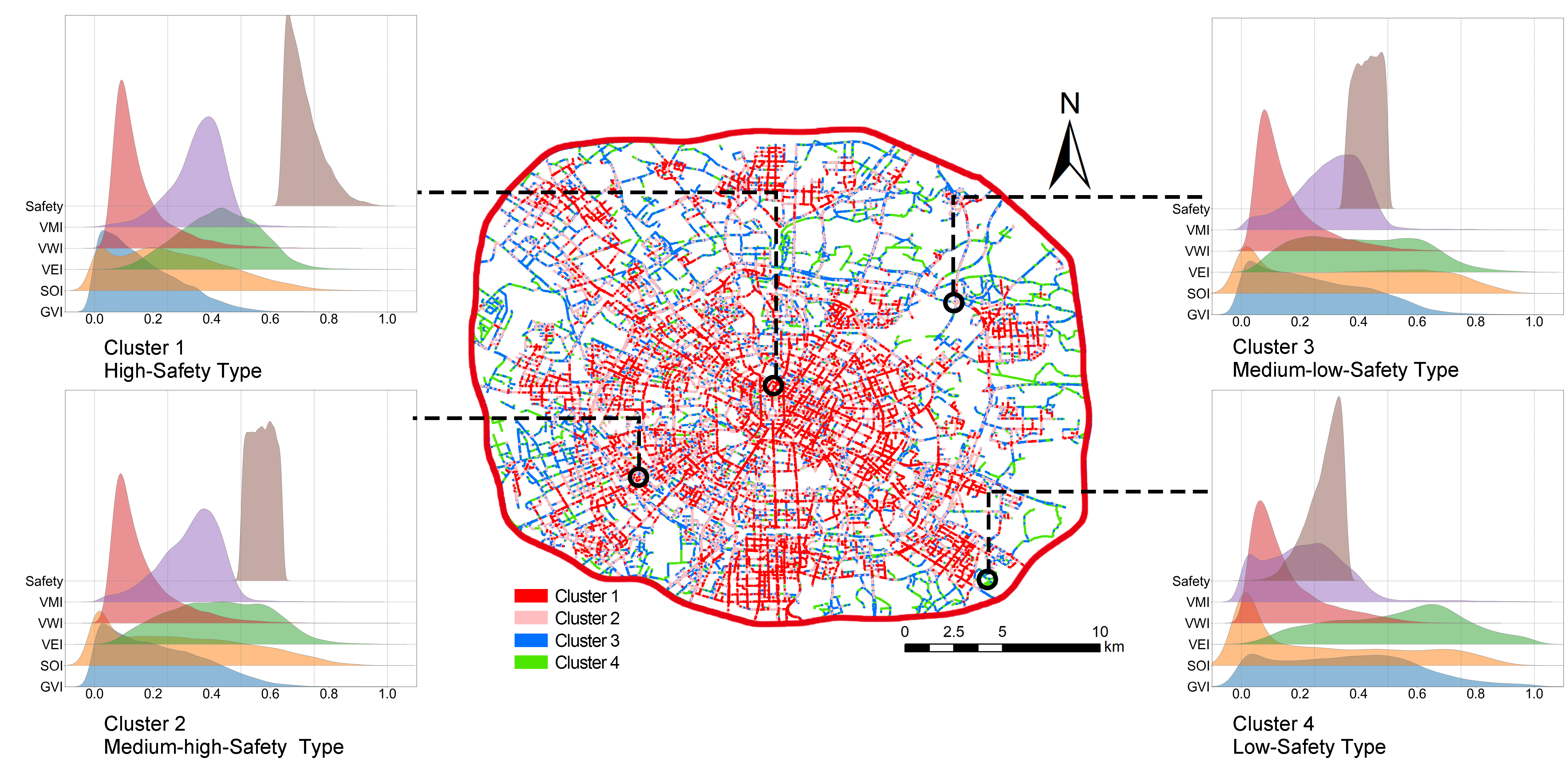}
\caption{Distribution of urban streets with different types of safety and representative cases.}
\label{fig:mapping}
\end{figure*}

Exploring the error propagation from the MLLM output through CLIP and ultimately to human annotations is essential. To examine this, we conducted experiments using LLaVA-NeXT, GPT-4V, and GPT-4o to generate the anchor set for continuous KNN+CLIP, respectively. As illustrated in Figure 7, the propagation effect—represented by the red bar as $K^2$ of KNN+CLIP results compared to human annotations—demonstrates a noticeable transfer of error from the MLLM output to CLIP and then to human annotations. This effect is more pronounced with MLLMs that exhibit lower alignment with human judgments, emphasizing the critical role of high-performing MLLMs. Furthermore, our observations reveal that anchor sets generated by higher-quality MLLMs yield improved KNN+CLIP results, especially prominent in the case of Chengdu city.

\subsection{Safety Perception Clustering at City Scale}

The results in Figure \ref{fig:mapping} indicate that urban safety perception in the central urban area of Chengdu can be categorized into four types: high-safety, medium-high-safety, medium-low-safety, and low-safety, accounting for 19.4\%, 33.3\%, 30.0\%, and 17.1\%, respectively. Using hierarchical clustering based on the local effects of SVI visual indicator variables, we can explore the impact of different physical elements' visual proportions on urban safety perception. Figure also shows the spatial distribution of these clusters and the density plots of their main visual indicator variables.

To better understand the factors influencing safety perception, we analyze key visual indicators related to the physical environment. These indicators, such as the Visual Enclosure Index (VEI), Visual Mobility Index (VMI), Green View Index (GVI), Sky Openness Index (SOI), and Visual Walkability Index (VWI), provide a quantitative basis for examining how different environmental elements impact perceptions of safety \cite{Yang2024}. The high-safety cluster is primarily located in the city center and areas with well-developed built environments, characterized by a high VEI and VMI but a low GVI. Dense buildings and heavy traffic increase human activity and surveillance, thereby providing a higher perception of safety. The medium-high-safety cluster is mainly distributed in the urban core and around major traffic routes. Although its VEI, VMI, and SOI are slightly lower than those of the high-safety cluster, it has a higher VWI and GVI. A good walking environment and moderate greening enhance residents' outdoor activity time and social interaction, thereby improving safety perception.

The medium-low-safety cluster is located on the periphery of the urban center, with moderate levels of VMI, VWI, GVI, and SOI. The scattered layout of buildings and roads, along with poor greening and walking conditions, may lead to a weaker safety perception in these areas. The low-safety cluster is mainly distributed in the outermost areas of the city, near the fourth ring road. This cluster features high SOI and GVI but low VEI. Low building density and reduced human activity, coupled with a lack of necessary surveillance and public activities, result in a decrease in safety perception.

\subsection{Nonlinear Associations of Factor Identification}

\begin{figure*}[t]
\centering
\includegraphics[width=1\textwidth]{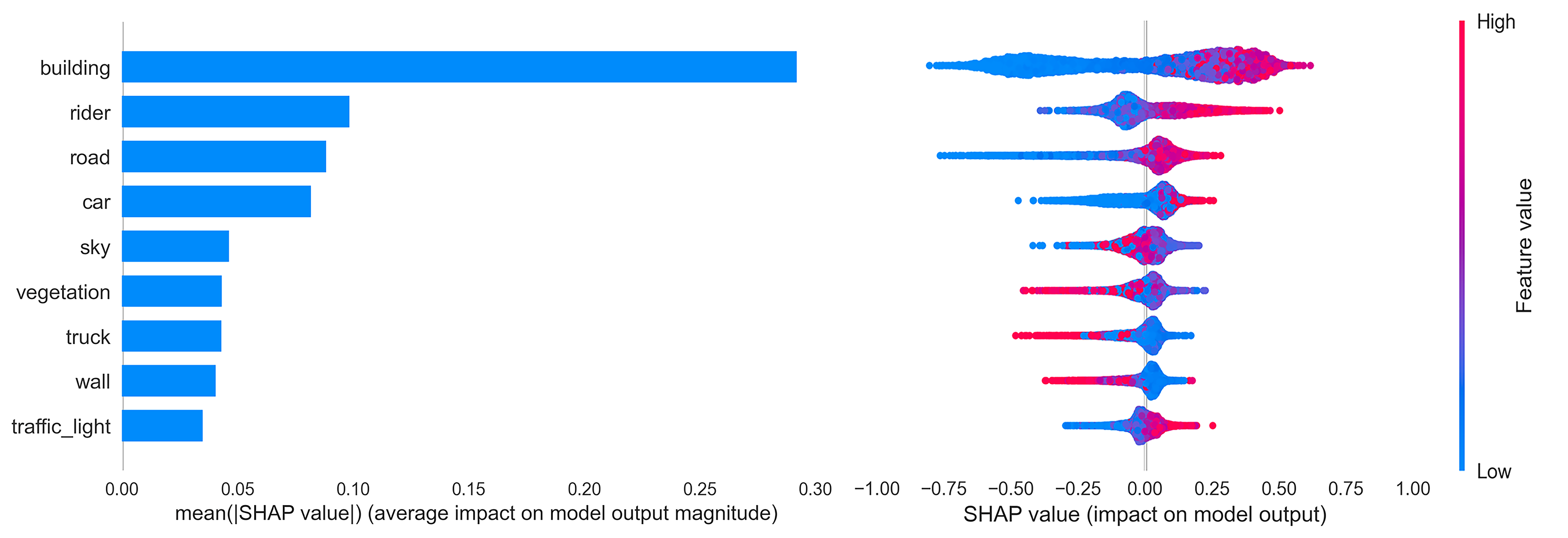}
\caption{Global and local relative importance diagram of SHAP model.}
\label{fig:relative}
\end{figure*}

By quantifying the contribution of each input variable to the prediction outcomes through SHapley Additive exPlanations (SHAP) values, we can gain a deeper understanding of whether the results of the prediction model, based on MLLM scoring, are consistent with human experiences of street safety. We utilized the eXtreme Gradient Boosting (XGBoost) model to establish the relationship between safety perception scores and 19 physical elements of the SVI, and employed SHAP values to interpret the model results. We utilized DeepLab v3 with a pre-trained model on the Cityscape dataset to perform semantic segmentation of 19 physical elements in the SVIs and calculated the proportion of each element in the images. Experimental results show that the XGBoost model can effectively capture the nonlinear relationship between segmentation results (such as the proportions of visual elements like sky, buildings, vegetation, etc.) and safety perception scores. The model achieved a Mean Squared Error (MSE) of 0.178 and an R² of 0.947 on the test set, indicating high predictive accuracy and stability.

\begin{figure*}[t]
    \centering
    \includegraphics[width=0.9\textwidth]{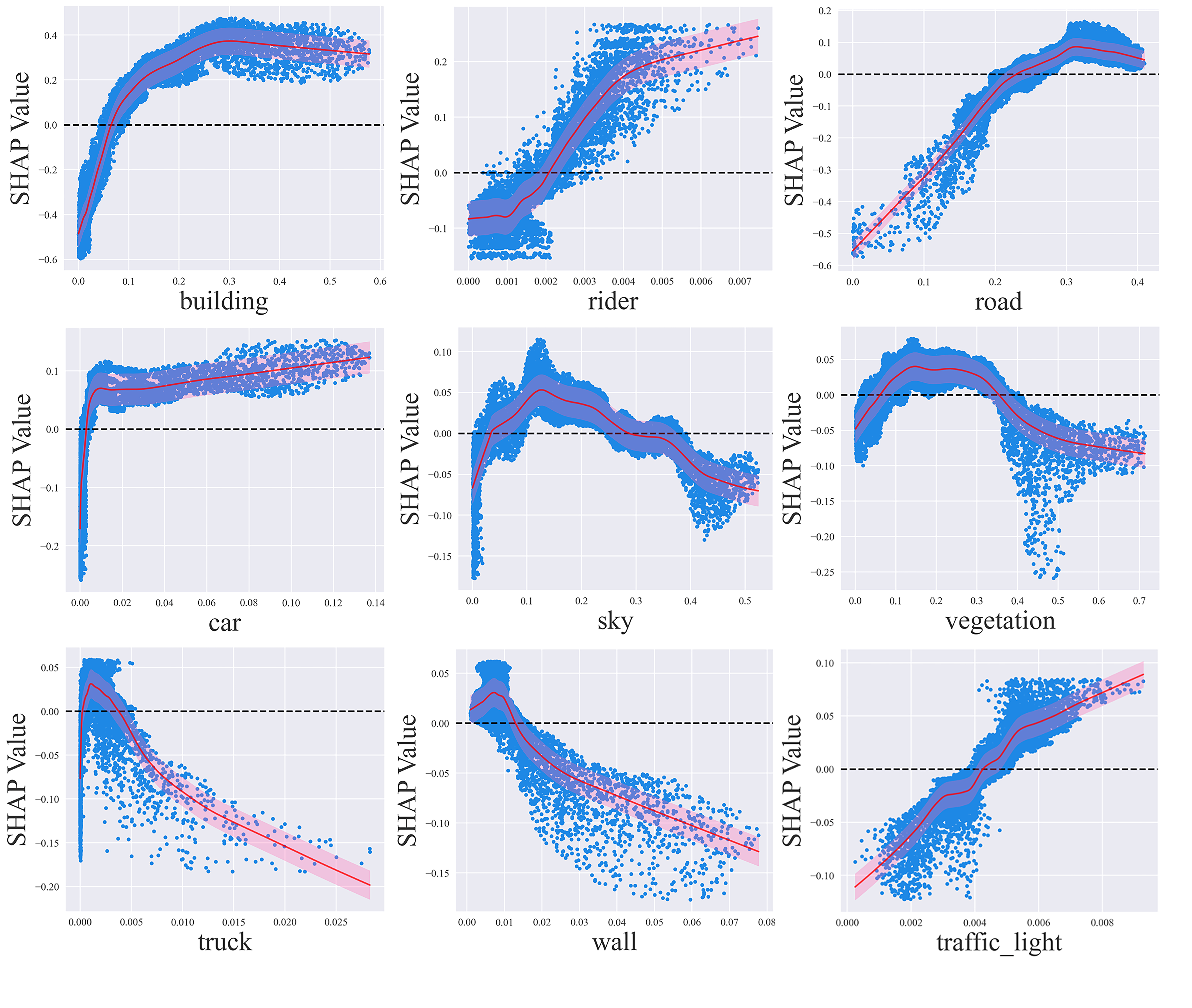}
    \caption{Local dependence diagram of urban multi-source data on safety perception.}
    \label{fig:dependence}
\end{figure*}

Figure \ref{fig:relative} illustrates the global importance and local explanation plots of various physical element variables. The variables are ranked in descending order of global importance, showcasing the top nine indicators. In the local explanation plots, red to blue dots represent feature values from high to low, respectively, with the x-axis SHAP values indicating the positive and negative impacts on safety perception. Overall, the three most significant contributing factors are buildings, roads, and cars, all of which positively influence safety perception. Partial dependence plots further visualize the SHAP values for each sample point, with the x-axis representing the normalized variable size and the y-axis representing SHAP values. We presented the top nine independent variables in terms of relative importance (Figure \ref{fig:dependence}). These plots help explore the nonlinear and threshold effects of individual independent variables on the dependent variable.

The results indicate that the proportion of different elements has significant nonlinear effects on safety perception. The proportions of building elements, roads, cars, and sky elements exhibit complex influences on safety perception. Lower proportions of building elements negatively impact safety perception, but a moderate increase in building density can enhance it. However, when the proportions of buildings, riders, and roads elements exceed a certain threshold, their positive impact on the perception of security will tend to level off and may even diminish. 

In contrast, the proportions of wall and truck elements are significantly negatively correlated with safety perception. While wall elements can provide boundaries and a sense of enclosure, enhancing area safety to some extent, excessive walls can lead to oppressive spaces and visual monotony, reducing the attractiveness of public spaces. The persistent negative correlation between truck element proportions and safety perception indicates that high flows of heavy vehicles can cause noise, pollution, and safety risks, diminishing the appeal and safety of residential environments. 

Moreover, increases in the proportions of rider and traffic light elements are positively correlated with enhanced safety perception. The positive correlation between traffic light proportions and enhanced safety perception highlights the importance of traffic management facilities in improving road safety and order. Although increased vegetation proportions are also positively correlated with enhanced safety perception, in areas with high vegetation proportions, the positive impact may decrease or even turn negative. Similarly, although increasing sky elements at lower proportions enhances the perception of safety by improving visibility, exceeding a certain threshold diminishes this effect—or may even reverse it—as excessive openness can make individuals feel exposed and vulnerable.

\section{Discussion}

\subsection{Automatic Ranking Via MLLMs}
The primary findings of this study are verifying the capability of MLLMs to effectively assess urban safety perception through SVIs. The main advantages of this technology include enhanced operational automation and efficiency, ensuring consistency and reproducibility of evaluation results, and the ability to scale to different cities and broader geographical areas. As more data accumulates, the precision of safety assessments and the performance of the model can continuously improve, enabling dynamic updates and iterative enhancements. However, there are some drawbacks to using MLLMs for urban safety assessment. Firstly, the training and output of the model could be influenced by data biases, which might inadvertently reflect these biases in the results, affecting the fairness of decision-making. The performance requirements for the model are also very high, for example, only the proprietary GPT-4o model has achieved satisfactory results. Secondly, an over-reliance on technology might overlook the value of local knowledge and human intuition in safety perception. Although the model can provide quantitative scores, it is often difficult to explain the specific reasons behind these scores, which limits its application in scenarios requiring high interpretability.

Additionally, Kang et al. \cite{kang2023assessing} emphasize the limitations of global models in capturing localized contexts, noting their potential to introduce perceptual biases when interpreting local data. While the objectivity of MLLMs is generally beneficial, it can inadvertently lead to biases in the perception of localized data, presenting a persistent challenge. To address this, we explored two potential approaches. The first approach leveraged the role-playing capabilities of MLLMs. For example, when analyzing data from Chengdu, the model was prompted to adopt the perspective of a local resident. However, in 100 experiments conducted with GPT-4o, this method did not produce any noticeable changes in outcomes, suggesting its limited effectiveness in mitigating bias. The second approach considered fine-tuning the model to better adapt to localized contexts. However, this strategy requires substantial labeled data and faces significant technical and logistical hurdles, especially when fine-tuning closed-source models like GPT-4o. While addressing model biases is a critical goal, overcoming these challenges remains a complex and unresolved issue at this stage.

\subsection{Overall Urban Analysis}
We developed an innovative and efficient methodology that enables global-scale assessments. This method specifically leverages a sampled anchor set to extrapolate safety metrics, effectively covering the entire urban landscape. Unlike previous methods, our approach obviates the need for a training phase, a significant advantage when data availability is limited. This innovative methodology not only enhances the precision of safety assessments over large regions but also significantly improves scalability. As a result, it can be adapted to a variety of urban settings around the world, providing a robust tool for urban planners and policymakers to assess and enhance public safety effectively. One notable limitation of our methodology lies in its reliance on the pre-trained CLIP model, which presents challenges when safety visual indicators diverge significantly from the data used to train the model. In such cases, the model’s capacity to accurately assess similarities may be reduced, potentially resulting in ambiguous outputs. Nonetheless, our findings suggest that this approach still outperforms traditional training-based methods, highlighting the substantial value of even an off-the-shelf pre-trained CLIP model in assessing urban safety. While a CLIP model fine-tuned specifically for urban safety analysis could likely yield even better results, the primary obstacle remains the current scarcity of relevant datasets for such fine-tuning. Addressing this data gap presents a valuable direction for future research, with the potential to enhance model accuracy and expand its applicability in urban safety evaluation.

\subsection{Potential Applications}
The application of MLLMs for automated safety perception assessment holds profound and extensive potential in urban environments. In the realm of urban planning and design, MLLMs can swiftly identify high-risk areas, enabling targeted safety interventions such as improved lighting and optimized public spaces. By providing quantitative insights into safety perceptions, MLLMs support data-driven decision-making, thereby justifying and informing urban development projects. In the field of social research and policy-making, MLLMs offer invaluable tools for understanding the complex relationships between perceived safety and various social factors, such as crime rates and socioeconomic conditions. This technology allows for rigorous evaluation of policy impacts by comparing safety perception data before and after the implementation of urban improvement initiatives. Additionally, by incorporating residents' feedback into safety assessments, MLLMs foster community engagement and participatory governance. The integration of MLLMs also enhances human-machine collaboration by augmenting human capabilities in processing and analyzing large datasets, ensuring consistent and objective evaluations. This collaboration enables urban planners, researchers, and policymakers to focus on strategic decision-making while leveraging the computational power of MLLMs to derive actionable insights. Given the performance of MLLMs in substituting human annotation for urban safety perception tasks, they also hold the potential for automating other urban perception tasks.

In our work, we also emphasize the importance of ensuring that MLLMs not only provide answers but also explain their reasoning. For example, as shown in Table \ref{table_sample_gpt4}, when comparing the SVIs, we prompt MLLMs to justify their responses. This approach transforms the outputs from mere black-box predictions into interpretable decisions, fostering trust and accountability in their application. However, the interpretability of MLLMs remains a significant challenge in the field. Although our methodology incorporates explicit prompts for generating explanations, the resulting rationales are often difficult to evaluate systematically. This limitation not only impacts usability but also raises concerns about safety in real-world applications. In particular, incomplete or misleading explanations could lead to misinformed decisions in safety assessments. Therefore, further advancements are needed to enhance MLLMs’ ability to generate explanations that are not only transparent but also accurate and contextually reliable.

\section{Conclusion}
In this paper, we introduce an innovative automatic pipeline designed to evaluate perceptions of urban safety. Utilizing advanced MLLMs, our approach enables an automated safety ranking that eliminates the need for manual human annotation while maintaining a high correlation with human perception. Furthermore, we have developed a retrieval-based technique that allows for the efficient evaluation of SVIs across expansive urban environments. Our findings reveal significant correlations between the computed safety scores and various urban elements, suggesting that our model effectively captures the essential aspects of perceived safety. Such advancements hold promise for facilitating more adaptive and data-informed urban management strategies, potentially transforming the landscape of urban governance and planning. Our future plan aims to expand the capabilities of our pipeline by incorporating real-time data analysis to dynamically update safety perceptions based on changing urban conditions. We intend to integrate diverse data sources, including social media feeds, traffic patterns, and public surveillance footage, to enhance the accuracy and relevance of our safety assessments.

\section*{Acknowledgment}
This research was supported by the General Project (Youth) of Humanities and Social Sciences in Jiangxi Province, China (JC24203), and by the Natural Science Foundation of Jiangxi Province, China (20242BAB20223). This work was also supported by JST ACT-X (JPMJAX24C8) and the World Premier International Research Center Initiative (WPI), MEXT, Japan.

\bibliographystyle{elsarticle-num} 
\bibliography{example}

\begin{thebibliography}{10}
\expandafter\ifx\csname url\endcsname\relax
  \def\url#1{\texttt{#1}}\fi
\expandafter\ifx\csname urlprefix\endcsname\relax\def\urlprefix{URL }\fi
\expandafter\ifx\csname href\endcsname\relax
  \def\href#1#2{#2} \def\path#1{#1}\fi

\bibitem{pol2019safe}
P.~Pol, The Safe City: Safety and Urban Development in European Cities, Routledge, New York, 2019.

\bibitem{zhao2023}
P.~Zhao, L.~Zeng, Transport safety, in: Transport Efficiency and Safety in China, Springer Nature Singapore, Singapore, 2023, pp. 313--343.

\bibitem{qiu2023subjective}
W.~Qiu, W.~Li, X.~Liu, Z.~Zhang, X.~Li, X.~Huang, Subjective and objective measures of streetscape perceptions: Relationships with property value in shanghai, Cities 132 (2023) 104037.

\bibitem{wang2022measuring}
L.~Wang, X.~Han, J.~He, T.~Jung, Measuring residents’ perceptions of city streets to inform better street planning through deep learning and space syntax, ISPRS Journal of Photogrammetry and Remote Sensing 190 (2022) 215--230.

\bibitem{xue2020extracting}
Y.~Xue, C.~Li, Extracting chinese geographic data from baidu map api, The Stata Journal 20~(4) (2020) 805--811.

\bibitem{fan2023urban}
Z.~Fan, F.~Zhang, B.~P. Loo, C.~Ratti, Urban visual intelligence: Uncovering hidden city profiles with street view images, Proceedings of the National Academy of Sciences 120~(27) (2023) e2220417120.

\bibitem{yao2019human}
Y.~Yao, Z.~Liang, Z.~Yuan, P.~Liu, Y.~Bie, J.~Zhang, R.~Wang, J.~Wang, Q.~Guan, A human-machine adversarial scoring framework for urban perception assessment using street-view images, International Journal of Geographical Information Science 33~(12) (2019) 2363--2384.

\bibitem{salesses2012place}
M.~P. Salesses, \href{https://dspace.mit.edu/handle/1721.1/76533}{Place pulse: Measuring the collaborative image of the city}, Masters thesis, Massachusetts Institute of Technology, Cambridge, MA (2012).
\newline\urlprefix\url{https://dspace.mit.edu/handle/1721.1/76533}

\bibitem{salesses2013collaborative}
P.~Salesses, K.~Schechtner, C.~A. Hidalgo, The collaborative image of the city: mapping the inequality of urban perception, PloS one 8~(7) (2013) e68400.

\bibitem{zhang2018measuring}
F.~Zhang, B.~Zhou, L.~Liu, Y.~Liu, H.~H. Fung, H.~Lin, C.~Ratti, Measuring human perceptions of a large-scale urban region using machine learning, Landscape and Urban Planning 180 (2018) 148--160.

\bibitem{LLMSurvey}
W.~X. Zhao, K.~Zhou, J.~Li, T.~Tang, X.~Wang, Y.~Hou, Y.~Min, B.~Zhang, J.~Zhang, Z.~Dong, Y.~Du, C.~Yang, Y.~Chen, Z.~Chen, J.~Jiang, R.~Ren, Y.~Li, X.~Tang, Z.~Liu, P.~Liu, J.-Y. Nie, J.-R. Wen, \href{http://arxiv.org/abs/2303.18223}{A survey of large language models}, arXiv preprint arXiv:2303.18223 (2023).
\newline\urlprefix\url{http://arxiv.org/abs/2303.18223}

\bibitem{yin2023survey}
S.~Yin, C.~Fu, S.~Zhao, K.~Li, X.~Sun, T.~Xu, E.~Chen, A survey on multimodal large language models, arXiv preprint arXiv:2306.13549 (2023).

\bibitem{gpt4}
OpenAI, Gpt-4 technical report, arXiv (2023).

\bibitem{li2023blip}
J.~Li, D.~Li, S.~Savarese, S.~Hoi, Blip-2: Bootstrapping language-image pre-training with frozen image encoders and large language models, arXiv preprint arXiv:2301.12597 (2023).

\bibitem{wang2023botcl}
B.~Wang, L.~Li, Y.~Nakashima, H.~Nagahara, Learning bottleneck concepts in image classification, in: IEEE Conference on Computer Vision and Pattern Recognition (CVPR), 2023, pp. 10962--10971.

\bibitem{liu2024visual}
H.~Liu, C.~Li, Q.~Wu, Y.~J. Lee, Visual instruction tuning, Advances in neural information processing systems 36 (2024).

\bibitem{zhu2023minigpt}
D.~Zhu, J.~Chen, X.~Shen, X.~Li, M.~Elhoseiny, Minigpt-4: Enhancing vision-language understanding with advanced large language models, arXiv preprint arXiv:2304.10592 (2023).

\bibitem{wang2023match}
B.~Wang, L.~Li, M.~Verma, Y.~Nakashima, R.~Kawasaki, H.~Nagahara, Match them up: visually explainable few-shot image classification, Applied Intelligence 53~(9) (2023) 10956--10977.

\bibitem{radford2021learning}
A.~Radford, J.~W. Kim, C.~Hallacy, A.~Ramesh, G.~Goh, S.~Agarwal, G.~Sastry, A.~Askell, P.~Mishkin, J.~Clark, et~al., Learning transferable visual models from natural language supervision, in: International conference on machine learning, PMLR, 2021, pp. 8748--8763.

\bibitem{gehl2011life}
J.~Gehl, Life Between Buildings: Using Public Space, Island Press, 2011.

\bibitem{yoshinobu1986aesthetic}
Y.~Ashihara, The Aesthetic Townscape, MIT Press, 1986.

\bibitem{jacobs1961death}
J.~Jacobs, The Death and Life of Great American Cities, Random House, 1961.

\bibitem{Danish2024}
M.~Danish, S.~Labib, B.~Ricker, M.~Helbich, A citizen science toolkit to collect human perceptions of urban environments using open street view images (2024).
\newblock \href {http://arxiv.org/abs/2403.00174} {\path{arXiv:2403.00174}}, \href {https://doi.org/10.48550/arXiv.2403.00174} {\path{doi:10.48550/arXiv.2403.00174}}.

\bibitem{mehta2014evaluating}
V.~Mehta, Evaluating public space, Journal of Urban design 19~(1) (2014) 53--88.

\bibitem{mouratidis2019impact}
K.~Mouratidis, The impact of urban tree cover on perceived safety, Urban Forestry \& Urban Greening 44 (2019) 126434.

\bibitem{zeng2022perceived}
E.~Zeng, Y.~Dong, L.~Yan, A.~Lin, Perceived safety in the neighborhood: Exploring the role of built environment, social factors, physical activity and multiple pathways of influence, Buildings 13~(1) (2022) 2.

\bibitem{makinde2020correlates}
O.~O. Makinde, The correlates of residents’ perception of safety in gated communities in nigeria, Social Sciences \& Humanities Open 2~(1) (2020) 100018.

\bibitem{zhang2021perception}
F.~Zhang, Z.~Fan, Y.~Kang, Y.~Hu, C.~Ratti, “perception bias”: Deciphering a mismatch between urban crime and perception of safety, Landscape and Urban Planning 207 (2021) 104003.

\bibitem{kang2023assessing}
Y.~Kang, J.~Abraham, V.~Ceccato, F.~Duarte, S.~Gao, L.~Ljungqvist, F.~Zhang, P.~N{\"a}sman, C.~Ratti, Assessing differences in safety perceptions using geoai and survey across neighbourhoods in stockholm, sweden, Landscape and Urban Planning 236 (2023) 104768.

\bibitem{cui2023analysing}
Q.~Cui, Y.~Zhang, G.~Yang, Y.~Huang, Y.~Chen, Analysing gender differences in the perceived safety from street view imagery, International Journal of Applied Earth Observation and Geoinformation 124 (2023) 103537.

\bibitem{liu2023interpretable}
Y.~Liu, M.~Chen, M.~Wang, J.~Huang, F.~Thomas, K.~Rahimi, M.~Mamouei, An interpretable machine learning framework for measuring urban perceptions from panoramic street view images, Iscience 26~(3) (2023).

\bibitem{keizer2008spreading}
K.~Keizer, S.~Lindenberg, L.~Steg, The spreading of disorder, science 322~(5908) (2008) 1681--1685.

\bibitem{kelling1982broken}
G.~L. Kelling, J.~Q. Wilson, et~al., Broken windows, Atlantic monthly 249~(3) (1982) 29--38.

\bibitem{nasar1990evaluative}
J.~L. Nasar, The evaluative image of the city, Journal of the American Planning Association 56~(1) (1990) 41--53.

\bibitem{halpern2014mental}
D.~Halpern, Mental health and the built environment: more than bricks and mortar?, Routledge, 2014.

\bibitem{wang2024improving}
B.~Wang, J.~Zhang, R.~Zhang, Y.~Li, L.~Li, Y.~Nakashima, Improving facade parsing with vision transformers and line integration, Advanced Engineering Informatics 60 (2024) 102463.

\bibitem{griew2013developing}
P.~Griew, M.~Hillsdon, C.~Foster, E.~Coombes, A.~Jones, P.~Wilkinson, Developing and testing a street audit tool using google street view to measure environmental supportiveness for physical activity, International Journal of Behavioral Nutrition and Physical Activity 10 (2013) 1--7.

\bibitem{naik2014streetscore}
N.~Naik, J.~Philipoom, R.~Raskar, C.~Hidalgo, Streetscore-predicting the perceived safety of one million streetscapes, in: Proceedings of the IEEE conference on computer vision and pattern recognition workshops, 2014, pp. 779--785.

\bibitem{naik2016cities}
N.~Naik, R.~Raskar, C.~A. Hidalgo, Cities are physical too: Using computer vision to measure the quality and impact of urban appearance, American Economic Review 106~(5) (2016) 128--132.

\bibitem{krizhevsky2012imagenet}
A.~Krizhevsky, I.~Sutskever, G.~E. Hinton, Imagenet classification with deep convolutional neural networks, Advances in neural information processing systems 25 (2012).

\bibitem{porzi2015predicting}
L.~Porzi, S.~Rota~Bul{\`o}, B.~Lepri, E.~Ricci, Predicting and understanding urban perception with convolutional neural networks, in: Proceedings of the 23rd ACM international conference on Multimedia, 2015, pp. 139--148.

\bibitem{russakovsky2015imagenet}
O.~Russakovsky, J.~Deng, H.~Su, J.~Krause, S.~Satheesh, S.~Ma, Z.~Huang, A.~Karpathy, A.~Khosla, M.~Bernstein, et~al., Imagenet large scale visual recognition challenge, International journal of computer vision 115 (2015) 211--252.

\bibitem{dubey2016deep}
A.~Dubey, N.~Naik, D.~Parikh, R.~Raskar, C.~A. Hidalgo, Deep learning the city: Quantifying urban perception at a global scale, in: Computer Vision--ECCV 2016: 14th European Conference, Amsterdam, The Netherlands, October 11--14, 2016, Proceedings, Part I 14, Springer, 2016, pp. 196--212.

\bibitem{naik2017computer}
N.~Naik, S.~D. Kominers, R.~Raskar, E.~L. Glaeser, C.~A. Hidalgo, Computer vision uncovers predictors of physical urban change, Proceedings of the National Academy of Sciences 114~(29) (2017) 7571--7576.

\bibitem{zhao2022}
P.~Zhao, Y.~Gao, Public transit travel choice in the post covid-19 pandemic era: An application of the extended theory of planned behavior, Travel Behaviour and Society 28 (2022) 181--195.

\bibitem{li2022measuring}
Y.~Li, N.~Yabuki, T.~Fukuda, Measuring visual walkability perception using panoramic street view images, virtual reality, and deep learning, Sustainable Cities and Society 86 (2022) 104140.

\bibitem{liu2024day}
Z.~Liu, T.~Li, T.~Ren, D.~Chen, W.~Li, W.~Qiu, Day-to-night street view image generation for 24-hour urban scene auditing using generative ai, Journal of Imaging 10~(5) (2024) 112.

\bibitem{lu2024using}
Y.~Lu, H.-M. Chen, Using google street view to reveal environmental justice: Assessing public perceived walkability in macroscale city, Landscape and Urban Planning 244 (2024) 104995.

\bibitem{wei2024measuring}
Z.~Wei, K.~Cao, M.-P. Kwan, Y.~Jiang, Q.~Feng, Measuring the age-friendliness of streets' walking environment using multi-source big data: A case study in shanghai, china, Cities 148 (2024) 104829.

\bibitem{bahrini2023chatgpt}
A.~Bahrini, M.~Khamoshifar, H.~Abbasimehr, R.~J. Riggs, M.~Esmaeili, R.~M. Majdabadkohne, M.~Pasehvar, Chatgpt: Applications, opportunities, and threats, in: 2023 Systems and Information Engineering Design Symposium (SIEDS), IEEE, 2023, pp. 274--279.

\bibitem{achiam2023gpt}
J.~Achiam, S.~Adler, S.~Agarwal, L.~Ahmad, I.~Akkaya, F.~L. Aleman, D.~Almeida, J.~Altenschmidt, S.~Altman, S.~Anadkat, et~al., Gpt-4 technical report, arXiv preprint arXiv:2303.08774 (2023).

\bibitem{chowdhery2023palm}
A.~Chowdhery, S.~Narang, J.~Devlin, M.~Bosma, G.~Mishra, A.~Roberts, P.~Barham, H.~W. Chung, C.~Sutton, S.~Gehrmann, et~al., Palm: Scaling language modeling with pathways, Journal of Machine Learning Research 24~(240) (2023) 1--113.

\bibitem{touvron2023llama}
H.~Touvron, T.~Lavril, G.~Izacard, X.~Martinet, M.-A. Lachaux, T.~Lacroix, B.~Rozi{\`e}re, N.~Goyal, E.~Hambro, F.~Azhar, et~al., Llama: Open and efficient foundation language models, arXiv preprint arXiv:2302.13971 (2023).

\bibitem{zhang2023knowledge}
Y.~Zhang, P.~Liu, F.~Biljecki, Knowledge and topology: A two layer spatially dependent graph neural networks to identify urban functions with time-series street view image, ISPRS Journal of Photogrammetry and Remote Sensing 198 (2023) 153--168.

\bibitem{herbrich2006trueskill}
R.~Herbrich, T.~Minka, T.~Graepel, Trueskill™: a bayesian skill rating system, Advances in neural information processing systems 19 (2006).

\bibitem{louviere2015best}
J.~J. Louviere, T.~N. Flynn, A.~A.~J. Marley, Best-worst scaling: Theory, methods and applications, Cambridge University Press, 2015.

\bibitem{liu2023tcra}
J.~Liu, L.~Li, T.~Xiang, B.~Wang, Y.~Qian, Tcra-llm: Token compression retrieval augmented large language model for inference cost reduction, arXiv preprint arXiv:2310.15556 (2023).

\bibitem{liu2024llavanext}
H.~Liu, C.~Li, Y.~Li, B.~Li, Y.~Zhang, S.~Shen, Y.~J. Lee, \href{https://llava-vl.github.io/blog/2024-01-30-llava-next/}{Llava-next: Improved reasoning, ocr, and world knowledge} (2024).
\newline\urlprefix\url{https://llava-vl.github.io/blog/2024-01-30-llava-next/}

\bibitem{abdin2024phi}
M.~Abdin, S.~A. Jacobs, A.~A. Awan, J.~Aneja, A.~Awadallah, H.~Awadalla, N.~Bach, A.~Bahree, A.~Bakhtiari, H.~Behl, et~al., Phi-3 technical report: A highly capable language model locally on your phone, arXiv preprint arXiv:2404.14219 (2024).

\bibitem{zhang2023llama}
R.~Zhang, J.~Han, A.~Zhou, X.~Hu, S.~Yan, P.~Lu, H.~Li, P.~Gao, Y.~Qiao, Llama-adapter: Efficient fine-tuning of language models with zero-init attention, arXiv preprint arXiv:2303.16199 (2023).

\bibitem{dai2024instructblip}
W.~Dai, J.~Li, D.~Li, A.~M.~H. Tiong, J.~Zhao, W.~Wang, B.~Li, P.~N. Fung, S.~Hoi, Instructblip: Towards general-purpose vision-language models with instruction tuning, Advances in Neural Information Processing Systems 36 (2024).

\bibitem{mcinnes2018umap}
L.~McInnes, J.~Healy, J.~Melville, Umap: Uniform manifold approximation and projection for dimension reduction, arXiv preprint arXiv:1802.03426 (2018).

\bibitem{dai2021adaptive}
W.~Dai, X.~Li, W.~H.~K. Chiu, M.~D. Kuo, K.-T. Cheng, Adaptive contrast for image regression in computer-aided disease assessment, IEEE Transactions on Medical Imaging 41~(5) (2021) 1255--1268.

\bibitem{dai2023semi}
W.~Dai, X.~Li, K.-T. Cheng, Semi-supervised deep regression with uncertainty consistency and variational model ensembling via bayesian neural networks, in: Proceedings of the AAAI Conference on Artificial Intelligence, Vol.~37, 2023, pp. 7304--7313.

\bibitem{RESupCon}
Z.~Zhou, Y.~Zhao, H.~Zuo, W.~Chen, Ranking enhanced supervised contrastive learning for regression, in: Advances in Knowledge Discovery and Data Mining, 2024, pp. 15--27.

\bibitem{Yang2024}
W.~Yang, Y.~Li, Y.~Liu, P.~Fan, W.~Yue, Environmental factors for outdoor jogging in beijing: Insights from using explainable spatial machine learning and massive trajectory data, Landscape and Urban Planning 243 (2024) 104969.
\newblock \href {https://doi.org/10.1016/j.landurbplan.2023.104969} {\path{doi:10.1016/j.landurbplan.2023.104969}}.

\end{thebibliography}

% \bibliographystyle{elsarticle-harv} 
% \bibliography{example}

% % \bibitem[Author(year)]{label}
% % For example:
% \bibitem[1]{anisetti2017semi} Anisetti, M., Ardagna, C.A., Damiani, E., Gaudenzi, F., 2017. A semiautomatic and trustworthy scheme for continuous cloud service certification. IEEE Transactions on Services Computing 13, 30–43.

% \end{thebibliography}

\end{document}